\documentclass[10pt]{article}
\usepackage{amsthm,amssymb}

\usepackage{amsmath,amsfonts,latexsym,etoolbox,stmaryrd,graphicx}
\usepackage[noadjust]{cite}
\usepackage{pict2e}
\usepackage[shortcuts]{extdash}

\allowdisplaybreaks[4]

\newtoggle{CONF}
\newtoggle{arXiv}
\newtoggle{TR}
\newtoggle{false}
\newtoggle{true}\toggletrue{true}

\toggletrue{arXiv}
\toggletrue{TR}

\usepackage{environ}
\NewEnviron{wideformula}{%
  \iftoggle{CONF}{\begin{equation}\BODY\end{equation}}%
     {\begin{multline}\BODY\end{multline}}}
\NewEnviron{wideformula*}{%
  \iftoggle{CONF}{\begin{equation*}\BODY\end{equation*}}%
    {\begin{multline*}\BODY\end{multline*}}}

\NewEnviron{widish}{\iftoggle{CONF}{$\BODY$ }{\[\BODY\]}}
\NewEnviron{widish.}{\iftoggle{CONF}{$\BODY$. }{\[\BODY.\]}}

\usepackage[pdfpagemode=UseNone,pdfstartview=FitH]{hyperref}

\usepackage[T2A]{fontenc}
\newenvironment{cyr}{}{}

\usepackage{color}

\newcommand{\red}[1]{\textcolor{red}{#1}}

\newtheorem{theorem}{Theorem}

\newtheorem{corollary}{Corollary}

\theoremstyle{definition}

\theoremstyle{remark}
\newtheorem{remark}{Remark}

\NewDocumentEnvironment{myproof}{o}
  {%
    \IfNoValueTF{#1}
      {\begin{proof}}
      {%
        \iftoggle{CONF}
          {\begin{proof}[#1]}%
          {\begin{proof}[Proof {#1}]}
      }%
  }
  {%
    \iftoggle{CONF}
      {\qed\end{proof}}%
      {\end{proof}}
  }%
\newcommand{\N}{\mathbb{N}}   % the natural numbers
\newcommand{\Z}{\mathbb{Z}}   % the integer numbers
\newcommand{\Q}{\mathbb{Q}}   % the rational numbers
\newcommand{\dd}{\,\mathrm{d}}  % differential
\renewcommand{\d}{\mathrm{d}}   % differential for use in parentheses

\newcommand{\e}{\mathrm{e}}    % mathematical constant \approx 2.72

\renewcommand{\P}{\mathbb{P}}  % probability
\newcommand{\E}{\mathbb{E}}    % expectation

\DeclareMathOperator{\ER}{\mathcal{E}^{\mathrm{R}}}  % IID e-variables
\DeclareMathOperator{\EX}{\mathcal{E}^{\mathrm{X}}}  % exchangeability e-variables
\DeclareMathOperator{\PR}{\mathcal{P}^{\mathrm{R}}}  % IID p-variables
\DeclareMathOperator{\PX}{\mathcal{P}^{\mathrm{X}}}  % exchangeability p-variables
\DeclareMathOperator{\EtR}{\mathcal{E}^{\mathrm{tR}}} % train-invariant IID e-variables
\DeclareMathOperator{\EtX}{\mathcal{E}^{\mathrm{tX}}} % train-invariant exchangeability e-variables
\DeclareMathOperator{\PtR}{\mathcal{P}^{\mathrm{tR}}} % train-invariant IID p-variables
\DeclareMathOperator{\PtX}{\mathcal{P}^{\mathrm{tX}}} % train-invariant exchangeability p-variables
\DeclareMathOperator{\EiR}{\mathcal{E}^{\mathrm{iR}}} % invariant IID e-variables
\DeclareMathOperator{\PiR}{\mathcal{P}^{\mathrm{iR}}} % invariant IID p-variables

\newcommand{\EEE}{\mathcal{E}} % generic class

\DeclareMathOperator{\lb}{lb}       % binary logarithm

\DeclareMathOperator{\DeR}{\mathit{D}^{\mathrm{eR}}} % IID e-deficiency
\DeclareMathOperator{\DeX}{\mathit{D}^{\mathrm{eX}}} % exchangeability e-deficiency
\DeclareMathOperator{\DpR}{\mathit{D}^{\mathrm{pR}}} % IID p-deficiency
\DeclareMathOperator{\DpX}{\mathit{D}^{\mathrm{pX}}} % exchangeability p-deficiency
\DeclareMathOperator{\DetR}{\mathit{D}^{\mathrm{etR}}} % train-invariant IID e-deficiency
\DeclareMathOperator{\DetX}{\mathit{D}^{\mathrm{etX}}} % train-invariant exchangeability e-deficiency
\DeclareMathOperator{\DptR}{\mathit{D}^{\mathrm{ptR}}} % train-invariant IID p-deficiency
\DeclareMathOperator{\DptX}{\mathit{D}^{\mathrm{ptX}}} % train-invariant exchangeability p-deficiency
\newcommand{\eqa}{\mathbin{=^+}}
\newcommand{\lea}{\mathbin{\le^+}}
\newcommand{\gea}{\mathbin{\ge^+}}

\title{Randomness, exchangeability, and conformal prediction}

\author{Vladimir Vovk}

\begin{document}
\maketitle
\begin{abstract}%
  This paper argues for a wider use of the functional theory of randomness,
  a modification of the algorithmic theory of randomness
  getting rid of unspecified additive constants.
  Both theories are useful for understanding relationships
  between the assumptions of IID data
  and data exchangeability.
  While the assumption of IID data is standard in machine learning,
  conformal prediction relies on data exchangeability.
  Nouretdinov, V'yugin, and Gammerman showed,
  using the language of the algorithmic theory of randomness,
  that conformal prediction is a universal method
  under the assumption of IID data.
  In this paper (written for the Alex Gammerman Festschrift)
  I will selectively review connections between exchangeability
  and the property of being IID,
  early history of conformal prediction,
  my encounters and collaboration with Alex and other interesting people,
  and a translation of Nouretdinov et al.'s results
  into the language of the functional theory of randomness,
  which moves it closer to practice.
  Namely, the translation says that every confidence predictor
  that is valid for IID data
  can be transformed to a conformal predictor
  without losing much in predictive efficiency.
  \iftoggle{arXiv}{%

     The version of this paper at \url{http://alrw.net} (Working Paper 42)
     is updated most often.%
  }{}%
\end{abstract}

\section{Introduction}
\label{sec:introduction}

The functional theory of randomness was proposed in
\cite{Vovk:2020Yuri} under the name of non-algorithmic theory of randomness.
The algorithmic theory of randomness was started by Kolmogorov in the 1960s
\cite{Kolmogorov:1968} and has been developed in numerous papers and books
(see, e.g., \cite{Shen/etal:2017book}).
It has been a powerful source of intuition,
but its weakness is the dependence on the choice
of a specific universal partial computable function,
which leads to the presence of unspecified additive (sometimes multiplicative) constants
in its mathematical results.
Kolmogorov \cite[Sect.~3]{Kolmogorov:1965} speculated
that for natural universal partial computable functions
the additive constants will be in hundreds rather than in tens of thousands of bits,
but this accuracy is very far from being sufficient in machine-learning and statistical applications
(an additive constant of $100$ in the definition of Kolmogorov complexity
leads to the astronomical multiplicative constant of $2^{100}$ in the corresponding p-value).

The way of dealing with unspecified constants proposed in \cite{Vovk:2020Yuri}
is to express statements of the algorithmic theory of randomness
as relations between various function classes.
It will be introduced in Sect.~\ref{sec:functional}.
In this paper we will call this approach the functional theory of randomness.
While it loses somewhat in intuitive simplicity,
it is closer to practical machine learning and statistics.

The main message of this paper is that the functional theory of randomness
can be useful in the foundations of machine learning in general
and conformal prediction in particular.
The most standard assumption in machine learning
is that the data are generated in the IID fashion
(are independent and identically distributed).
An \emph{a priori} weaker assumption is that of exchangeability,
although for infinite data sequences
being generated in the IID fashion and exchangeability
turn out to be essentially equivalent
by the celebrated de Finetti representation theorem.
The classical work on relations between the two assumptions,
being IID and being exchangeable,
will be the topic of Sect.~\ref{sec:RE}.

The word ``random'' is often used in two very different senses:
in the sense of statistical randomness referring to IID data
(as in the title of \cite{Vovk/etal:2022-local})
% and \cite{Vovk/etal:2005-local}
and in the sense of algorithmic randomness
(as in the title of \cite{Shen/etal:2017book}).
In this paper I will try not to use ``random'' and its derivative ``randomness'' often
(apart from the expressions ``algorithmic theory of randomness''
and ``functional theory of randomness'').
For the former sense,
I will usually replace derivatives of ``random''
by compounds containing ``IID'',
such as the \emph{IID assumption} for the assumption
that the data are generated in the IID fashion
(so that simply replacing ``IID'' by ``independent identically distributed''
becomes impossible, as in \cite{Vovk/etal:2009AOS}).
For the latter sense, I will often use the word ``typical''
and its derivative ``typicalness'',
which was approved by Kolmogorov
\cite[Appendix~2, footnote~1]{Shen/etal:2017book}
in its Russian form \begin{cyr}типичность\end{cyr}
and used in \cite{Kolmogorov/Uspensky:1987-local}
(in fact written by Uspensky \cite[Introduction]{Kolmogorov/Uspensky:1987-local}).

The following two sections, \ref{sec:Kolmogorov} and~\ref{sec:Alex},
also contain personal elements.
In Sect.~\ref{sec:Kolmogorov} I describe meeting Andrei Kolmogorov
and working under his supervision on the relation between IID and exchangeability.
In Kolmogorov's frequentist philosophy of probability,
the IID property was at the very basis of the notion of probability.
In Sect.~\ref{sec:Alex} I describe meeting Alex and then Vladimir Vapnik
about 10 years later.
Vapnik was a second person who impressed me
by his full-hearted acceptance of the IID assumption,
which quickly led to the development of conformal prediction.

In my work under Kolmogorov,
I realized that for finite data sequences
the difference between IID and exchangeability is important.
However, conformal prediction only uses exchangeability.
Can we improve on conformal prediction
by using the stronger IID assumption?
The topic of Sect.~\ref{sec:NVG} is the fundamental result
by Nouretdinov, V'yugin, and Gammerman saying that only limited improvement is possible.
The result is stated in terms of the algorithmic theory of randomness,
and this makes it very intuitive,
although the intuition is sometimes obscured by dense notation.

Despite Nouretdinov et al.'s result being fundamental,
it has a serious conceptual weakness,
shared by the rest of the algorithmic theory of randomness,
as discussed earlier.
The result involves unspecified constants,
and so, strictly speaking,
cannot have any practical implications.
In Sect.~\ref{sec:functional} I will state this and several related results
in terms of the functional theory of randomness.

In this paper $e$ may stand for an e-value.
Euler's number (the base of the natural logarithms) is $\e\approx2.72$.
The notation for binary logarithm is $\lb$
\cite[Sect.~10.1.2]{BSI-local:2010}.
No detailed knowledge of the algorithmic theory of randomness is assumed
on the part of the reader,
but knowing the basics would be useful.

\section{IID and exchangeability}
\label{sec:RE}

Modelling data as IID observations is an ancient notion.
Already Jacob Bernoulli \cite{Bernoulli:1713} was using the IID assumption
to state his weak law of large numbers.
As Glenn Shafer reminds us in \cite[Shafer's comment]{Gammerman/Vovk:2007},
the IID case has been central to probability and statistics ever since,
``but its inadequacy was always obvious,
and Leibniz made the point in his letters to Bernoulli:
the world is in constant flux;
causes do not remain constant, and so probabilities do not remain constant''.
There is no doubt the IID assumption is very limited.

The real question is whether the IID assumption is a good \emph{starting point}.
It can be fundamental without being all-encompassing;
many other, perhaps much more realistic, cases may reduce in some way to the IID case.
For example, when dealing with prediction or decision making,
is it a good strategy first to explore in detail what can be achieved
under the IID assumption and then to try and relax it?
Or is the assumption so limited that it is best to start elsewhere?
My views about this have been drifting over time,
and even now I am uncertain about it.

One relatively modest extension of the IID assumption
is the assumption of exchangeability;
for potentially infinite data sequences one may argue
that it is not an extension at all.
Philip Dawid says in \cite[Sect.~7]{Vovk/Shafer:2025},
``For so long,
and it's still true of 97\% of everything
done in statistics and machine learning and everything,
the fundamental assumption is basically
we have just a bag of exchangeable goodies.
And I thought that was just so limiting; how boring.
There's a big wide world beyond that.''
This has been my feeling as well since I started thinking about such things
(and among my colleagues,
Shafer's and Dawid's views on the philosophy of probability
are perhaps closest to mine).
However, I have also been impressed by the power of algorithms
developed under IID and exchangeability
and by many ingenuous ways of greatly relaxing these assumptions.

It is not clear who introduced exchangeability
(see \cite{Dale:1985}
for the complicated history),
but the most well-known theorem about exchangeability is de Finetti's
(generalized in later papers by other people),
which connects it with IID.
Let $\mathbf{Z}$ be an \emph{observation space}
(formally, a measurable space),
and suppose that we observe its elements
$z_i\in\mathbf{Z}$, $i=1,2,\dots$,
sequentially.
The IID assumption is that the observations $z_i$ are generated
from an \emph{IID probability measure} $Q^{\infty}$,
$Q$ being a probability measure on $\mathbf{Z}$.
The assumption of exchangeability is that
they are generated from an \emph{exchangeable} probability measure
on $\mathbf{Z}^{\infty}$,
i.e., a probability measure that is invariant w.r.\ to permutations
of finitely many observations.

According to de Finetti's theorem (see, e.g., \cite[Theorem~1.49]{Schervish:1995}),
for infinite sequences, the IID and exchangeability assumptions are equivalent.
Namely, each exchangeable probability measure $R$ on $\mathbf{Z}^{\infty}$
is a convex mixture of IID probability measures:
there exists a probability measure $\mu$
on the family $\mathfrak{P}(\mathbf{Z})$
of all probability measures on $\mathbf{Z}$
such that
\[
  R
  =
  \int_{\mathfrak{P}(\mathbf{Z})}
  Q^{\infty} \;
  \mu(\d Q).
\]
The theorem makes the weak assumption that $\mathbf{Z}$ is a standard Borel space
(and then $\mathfrak{P}(\mathbf{Z})$ is equipped
with the smallest $\sigma$-algebra
making all evaluation functionals measurable).

I find it intuitively compelling that the assumption
that the data are generated from a statistical model $M$
(i.e., a family probability measures)
is equivalent to the assumption that the data are generated
from the convex hull $\bar M$ of that statistical model.
However, there are people who do not share this intuition
(e.g., a friendly reviewer for \cite{Vovk/etal:2025PR-local}),
so let me try to make it more explicit.
The most basic way of testing a statistical model $M$
(``Cournot's principle'')
is to select in advance a \emph{critical region} $A$ of a small probability
under any probability measure in $M$
and reject $M$ if the actual data happens to be in $A$.
Since
\[
  \sup_{R\in M}R(A)
  =
  \sup_{R\in\bar M}R(A),
\]
rejecting $M$ and rejecting $\bar M$ is the same thing.
This conclusion is not affected if Cournot's principle
is replaced by more sophisticated ways of hypothesis testing,
such as using p-variables and e-variables,
to be discussed starting from the next section.
In particular, the IID and exchangeability assumptions are the same thing.
This is far from being true for finite sequences,
as will be discussed in detail in Sections \ref{sec:Kolmogorov}--\ref{sec:Alex}.

The IID picture is fundamental in the frequentist theory of probability and statistics,
at least as it was presented and developed by Richard von Mises
\cite{Mises:1919,Mises:1928-local}
and Andrei Kolmogorov \cite[Sect.~I.2]{Kolmogorov:1933},
who was following von Mises.
It is well known that Kolmogorov was the first
to put the mathematical theory of probability
on a firm axiomatic basis in his 1933 book \cite{Kolmogorov:1933}.
However, while the axioms of probability introduced in this book
slowly gained universal acceptance (see, e.g., \cite{Shafer/Vovk:2006-local}),
the way in which Kolmogorov proposed to connect his axioms with reality
\cite[Sect.~I.2]{Kolmogorov:1933}
was informal and has never become widely accepted.
According to Kolmogorov's frequentist Principle~A,
introduced in \cite[Sect.~I.2]{Kolmogorov:1933},
we can say that an event $A$ has a probability $\P(A)$
under a system of conditions $\mathfrak{S}$ if
\begin{quote}
  One can be practically certain that
  if the system of conditions $\mathfrak{S}$
  is repeated a large number of times, $n$,
  and the event $A$ occurs $m$ times,
  then the ratio $m/n$ will differ only slightly from $\P(A)$.
\end{quote}
(When quoting \cite[Sect.~I.2]{Kolmogorov:1933}
I am using the translation given in \cite[Sect.~5.2.1]{Shafer/Vovk:2006-local}.)
This principle,
which Kolmogorov traces to von Mises's ideas
\cite[Sect.~I.2, footnote~1]{Kolmogorov:1933},
gives us a way of measuring $\P(A)$.
Presumably the repetitions in Principle~A are independent,
and so IID observations are at the heart of frequentist probability.

\begin{remark}
  Kolmogorov's approach was not purely frequentist.
  Alongside his frequentist Principle~A
  he also had a non-frequentist Principle~B,
  namely Cournot's principle:
  \begin{quote}
    If $\P(A)$ is very small,
    then one can be practically certain that the event $A$
    will not occur on a single realization of the conditions $\mathfrak{S}$.
  \end{quote}
  Kolmogorov \cite[Sect.~I.2]{Kolmogorov:1933} postulated both principles,
  but it can be argued that Principle B makes Principle~A redundant
  \cite[Sect.~5.2]{Shafer/Vovk:2006-local}.
  In hindsight, all the books that I have co-authored so far
  can be traced back either to Kolmogorov's Principle~A \cite{Vovk/etal:2022-local}
  or Principle~B \cite{Shafer/Vovk:2001,Shafer/Vovk:2019}.
\end{remark}

For a long time after the publication of his book \cite{Kolmogorov:1933},
Kolmogorov talked about connections of his axioms
with reality informally,
believing that von Mises's approach,
which was based on a flawed definition of an individual infinite sequence of IID observations,
could not be cleanly applied to the real world.
(See, e.g., \cite{Kolmogorov:1956}.)
The breakthrough came when Kolmogorov visited India in 1962 \cite{Kolmogorov:1963}.
He realized that von Mises's picture can be made applicable to finite sequences
(albeit in a way that looks awkward to me,
no doubt in hindsight, in view of his later elegant algorithmic approach,
which was much more typical of Kolmogorov).

Soon afterwards Kolmogorov came up with his algorithmic theory of randomness
subsuming his theory developed in India.
In particular, he formalized what it means for a finite binary sequence
to be a typical IID sequence.
Since Kolmogorov was only dealing with binary sequences,
he referred to typical IID sequences as ``Bernoulli sequences''.
The key to his definition was a notion of algorithmic complexity
(``Kolmogorov complexity''),
and typicalness was defined as maximal complexity in a finite set.
His theory was described in the papers \cite{Kolmogorov:1965,Kolmogorov:1968,Kolmogorov:1983}.

Martin-L\"of \cite{Martin-Lof:1966} translated Kolmogorov's definition of typicalness
into a more standard statistical language defining universal p-values.
Later Levin and G\'acs modified Martin-L\"of's definition
in an important way, but I will talk about this in the next section.

\section{Meeting Kolmogorov; IID vs exchangeability for finite sequences}
\label{sec:Kolmogorov}

In 1980 Kolmogorov became Head of the Department of Mathematical Logic
at Moscow University,
and in the same year I became his student.
This happened after I attended his talk aimed at undergraduate students
and spoke to Alexei Semenov afterwards,
who in his role of the departmental scientific secretary
(\begin{cyr}ученый секретарь кафедры\end{cyr})
took care of the administrative side.
First I did the specialized part of a Soviet combined BSc/MSc degree programme
under Kolmogorov's supervision
(the specialized part covering the last three years of the 5-year degree programme),
and then I did a PhD under Kolmogorov and Semenov.

One of the problems that Kolmogorov had for me
was to quantify the qualitative (and intuitively obvious) statement
that Bernoulli sequences satisfy his frequentist definition
as given in \cite{Kolmogorov:1963}.
This was a difficult problem that did not look particularly appealing to me
(some results in this direction were obtained later
by Kolmogorov himself and his other student Eugene Asarin;
see \cite[Theorem~3]{Asarin:1988-local} for Kolmogorov's result
and \cite[Theorem 1]{Asarin:1987-local} for Asarin's).
Instead, I investigated the relation between IID and exchangeability
for finite sequences.

As I mentioned in the previous section,
Kolmogorov was only interested in finite sequences
in his work on the foundations of probability
and believed that infinite sequences,
being empirically non-existent,
are irrelevant when discussing connections
between the mathematical theory of probability and reality.
At one point during a walk to a train station
Kolmogorov told me that we can only see finite sequences around us,
but in the quote from Kolmogorov given in
\cite[Chap.~7, bottom of p.~57]{Angelopoulos/Bates:2023}
the word ``only'' is misplaced
(Kolmogorov could not see any infinite sequences,
and neither could I).

In my first journal paper \cite{Vovk:1986-local}
I explored Kolmogorov's definition of Bernoulli sequences
and argued that it was a formalization of a different kind of typicalness,
not typicalness under IID.
The term that I used, on Alexander Zvonkin's advice,
for Kolmogorov's Bernoulli sequences was von Mises's ``collectives'',
but in reality I meant exchangeability
(and I talk about exchangeability in the technical report \cite{Vovk:1986-local}
containing the proofs).
After that I introduced a definition of typicalness under IID
for binary sequences
and characterized the difference between the two definitions.
To state these results,
let me give the relevant (standard) definitions.

I will use the notion of an aggregate of constructive objects,
as in \cite[Sect.~1.0.6]{Uspensky/Semenov:1993}.
This is an infinitely countable set whose elements can be effectively numbered,
such as the set $\Z$ of all integer numbers
or the set $\{0,1\}^*$ of all finite binary sequences.
Let the observation space $\mathbf{Z}$
range over the finite non-empty subsets of a fixed aggregate of constructive objects.
In Kolmogorov's work in this area and in my work reported in this section,
$\mathbf{Z}=\{0,1\}$,
but let me give more general definitions for later use
(e.g., in the context of conformal prediction).
Let $\N$ be the set of natural numbers;
by default we do not include 0 in $\N$, so that $\N=\{1,2,\dots\}$.

A real-valued function $f$ defined on an aggregate of constructive objects
is \emph{lower semicomputable} if there is an algorithm that,
when fed with $v$ in the domain of $f$ and $t\in\Q$
($\Q$ being the set of rational numbers),
eventually stops if and only if $f(v)>r$.
Similarly, it is \emph{upper semicomputable}
if this condition holds with $f(v)>r$ replaced by $f(v)<r$.
(And computability is the conjunction of lower and upper semicomputability.)

Let me start from a definition equivalent to Kolmogorov's,
which is closest to conformal prediction.
A \emph{p-test for exchangeability}
is a nonnegative upper semicomputable function $P$
that takes as an input $\mathbf{Z}$, $N\in\N$,
and a sequence $z_1,\dots,z_N$ in $\mathbf{Z}^N$
and that satisfies, for all $\mathbf{Z}$, all $N$,
and all exchangeable probability measures $R$ on $\mathbf{Z}^N$,
\begin{equation}\label{eq:P}
  \forall\epsilon\in(0,1):
  R(P\le\epsilon)
  \le
  \epsilon
\end{equation}
(we usually omit mentioning $\mathbf{Z}$ and $N$
as arguments of $P$;
remember that $\mathbf{Z}$ always ranges over the finite subsets
of a fixed aggregate of constructive objects).
The requirement \eqref{eq:P} is usually expressed
by saying that $P$ is a \emph{p-variable}
(and its values are \emph{p-values}).
There exists a smallest, to within a constant factor,
p-test for exchangeability,
which is then called \emph{universal}.
Let us fix a universal p-test for exchangeability
and let $\DpX$ stands for its minus binary logarithm.
We call $\DpX(z_1,\dots,z_N)$
the \emph{exchangeability p-deficiency} of the sequence $(z_1,\dots,z_N)$.
We can also allow $P$ to depend on a \emph{condition} (an integer number)
(and then $P$ is required to be lower semicomputable as function
of all its arguments, including the condition);
the full notation is then $\DpX(z_1,\dots,z_N\mid k)$,
where $k$ is the condition.

The function $\DpX$ can be defined to some degree arbitrarily;
different choices of $\DpX$, however, will coincide to within an additive constant.
This makes results of the algorithmic theory of randomness not applicable in practice.
When we say that two functions (such as $\DpX$) that are only defined
to within an additive constant coincide,
we mean, of course, that their difference is bounded.
In general, when discussing relations (such as inequalities) between such functions,
we will always ignore additive constants.
Without loss of generality,
we will assume that the function $\DpX$,
and similar functions introduced later in this paper,
are integer-valued.

In the case $\mathbf{Z}=\{0,1\}$,
$\DpX(z_1,\dots,z_N)$ coincides with Kolmogorov's definition of Bernoulliness,
as follows from \cite[Proposition~11]{Vovk:2025ML-local}.
Kolmogorov's expression for ``$\DpX(z_1,\dots,z_N)\le m$''
was ``$z_1,\dots,z_N$ is $m$-Bernoulli''
(``\begin{cyr}$m$-бернуллиевская\end{cyr}'').

In a similar way, we can define
the \emph{IID p-deficiency} $\DpR(z_1,\dots,z_N)$ of a sequence $(z_1,\dots,z_N)$.
The only difference is that we replace p-tests for exchangeability
by \emph{p-tests for IID} defined
by letting $R$ in the definition \eqref{eq:P}
range over the IID probability measures $Q^N$,
$Q$ being a probability measure on $\mathbf{Z}$ generating one observation.
A universal p-test for IID also exists;
we let $\DpR$ stands for its minus binary logarithm
and call $\DpR(z_1,\dots,z_N)$
the \emph{IID p-deficiency} of $z_1,\dots,z_N$.
This is equivalent to my definition proposed in \cite{Vovk:1986-local}.

\begin{figure}[bt]
  \vspace{0.5cm}
  \begin{center}
    \unitlength 0.50mm
    \begin{picture}(50,50)(-10,-10)  % the size of the box and the coordinates of the bottom left corner
      \thicklines
      \put(34,40){\line(-1,0){28}} % the top connection (black and thick)
      \thinlines
      \put(34,0){\red{\line(-1,0){28}}}  % the bottom connection (red)
      \put(40,5){\red{\line(0,1){30}}}   % the right connection (calibration, red)
      \put(0,5){\red{\line(0,1){30}}}    % the left connection (calibration)
      % and now the notation for the deficiencies:
      \put(-6,35){\framebox(12,10)[cc]{$\DpX$}}  % top left corner
      \put(-6,-5){\framebox(12,10)[cc]{$\DeX$}}  % bottom left corner
      \put(34,-5){\framebox(12,10)[cc]{$\DeR$}}  % bottom right corner
      \put(34,35){\framebox(12,10)[cc]{$\DpR$}}  % top right corner
    \end{picture}
  \end{center}
  \vspace{-0.5cm}
  \caption{Connections between 4 deficiencies of typicalness:
    The connection between $\DpX$ and $\DpR$ (shown as thick black line)
    is established via the chain
    $\DpX$--$\DeX$--$\DeR$--$\DpR$ (shown as thin red lines).\label{fig:1986}}
\end{figure}

To establish connections between $\DpX$ and $\DpR$,
I followed a strategy that is standard in the algorithmic theory of randomness.
While hypothesis testing in classical statistics is based on p-values,
e-values are mathematically much more convenient and often serve as a useful tool.
Universal e-values were introduced in the algorithmic theory of randomness
by Levin and then simplified by G\'acs \cite{Gacs:1980}
(see also \cite{Vovk/Vyugin:1993}),
without using this expression,
and nowadays non-universal e-values are gaining popularity in statistics
(see, e.g., \cite{Grunwald/etal:2024,Ramdas/Wang:book,Vovk/Wang:2021}).
Therefore, the Martin-L\"of-style functions $\DpX$ and $\DpR$
are connected by connecting $\DpX$ with $\DeX$, then $\DeX$ with $\DeR$,
and finally $\DeR$ with $\DpR$,
where $\DeX$ and $\DeR$ are Levin-style analogues of $\DpX$ and $\DpR$,
to be defined momentarily.
(The connections are shown in Fig.~\ref{fig:1986}.)

An \emph{e-test for exchangeability}
is a nonnegative lower semicomputable function $E$
on the same domain as a p-test for exchangeability,
but it is required to satisfy
\begin{equation}\label{eq:E-test}
  \sum_{\zeta\in\mathbf{Z}^N}
  E(\zeta)
  R(\{\zeta\})
  \le
  1
\end{equation}
in place of \eqref{eq:P}
for all $\mathbf{Z}$, $N$, and exchangeable $R$.
Both upper semicomputability for $P$ (in \eqref{eq:P})
and lower semicomputability for $E$ (in \eqref{eq:E-test})
are natural requirements:
we reject the null hypothesis of exchangeability (or IID)
when a p-value is small (say, below some threshold such as 1\% or 5\%)
or an e-value is large,
and the decision to reject should be taken in finite time.
The condition \eqref{eq:E-test} means that $E$ is an \emph{e-variable},
and then its values are \emph{e-values}.
We fix a universal (this time meaning largest to within a constant factor)
e-test for exchangeability
and call its binary logarithm $\DeX$ \emph{exchangeability e-deficiency}.
Replacing exchangeable $R$ by $R:=Q^N$,
we get the definition of $\DeR$, \emph{IID e-deficiency}.

For any data sequence $z_1,\dots,z_N$,
let us define the IID e-deficiency of the corresponding \emph{configuration}
$\lbag z_1,\dots,z_N\rbag$,
i.e., the bag (multiset) of its elements, as
\begin{equation}\label{eq:configuration}
  \DeR(\lbag z_1,\dots,z_N\rbag)
  :=
  \min_{\pi}
  \DeR(z_{\pi(1)},\dots,z_{\pi(N)}),
\end{equation}
$\pi$ ranging over all permutations of $\{1,\dots,N\}$.
In other words, a bag is IID (compatible with the IID assumption)
if it can arise from an IID data sequence.
If $\zeta$ is a data sequence, we let $\lbag\zeta\rbag$ stand for the bag of its elements.

The following relation between IID and exchangeability
is stated in \cite[Theorem~1]{Vovk:1986-local} for $\mathbf{Z}=\{0,1\}$
and in \cite[Theorem~3]{Vovk/etal:1999} in general.
It uses ``$\eqa$'' to mean coincidence of two functions
to within an additive constant;
$\mathbf{Z}^+$ is the family of all non-empty finite sequences of observations.
Remember that $\mathbf{Z}$ varies over the finite non-empty subsets
of a fixed aggregate of constructive objects.

\begin{theorem}\label{thm:connection}
  Let $\zeta$ range over $\mathbf{Z}^+$ for a variable $\mathbf{Z}$.
  Then
  \begin{equation}\label{eq:1986}
    \DeR(\zeta)
    \eqa
    \DeR(\lbag\zeta\rbag)
    +
    \DeX(\zeta\mid\DeR(\lbag\zeta\rbag)).
  \end{equation}
\end{theorem}

Theorem~\ref{thm:connection} clarifies the relation
between exchangeability and IID in the case of finite sequences:
a sequence is IID if and only if it is exchangeable and its configuration is IID.
The difference between the deficiencies of IID and exchangeability
of a data sequence is, roughly,
the IID deficiency of its configuration;
the condition ``${}\mid\DeR(\lbag\zeta\rbag$'' in \eqref{eq:1986}
slightly obscures this, but \eqref{eq:1986} implies, e.g.,
\begin{equation*}
  \DeR(\lbag\zeta\rbag)
  +
  \DeX(\zeta)
  \lea
  \DeR(\zeta)
  \lea
  1.01\DeR(\lbag\zeta\rbag)
  +
  \DeX(\zeta)
\end{equation*}
(with ``$\lea$'' denoting the inequality to within an additive constant).

Another result that I obtained in \cite{Vovk:1986-local}
(Theorem~2)
was about how big the difference given by \eqref{eq:1986}
between Kolmogorov's and my definitions can be.
In the binary case considered in that paper,
the configuration $\lbag\zeta\rbag$ carries the same information
as the number of 1s in $\zeta$ given its length $N$.
Let $k$ be the number of 1s.
Then $\DeR(k)$ can be characterized as the typicalness deficiency of $k$
in its neighbourhood of size $\sqrt{k(N-k)/N}$
(approximately $\sqrt{N}$ if $k$ is neither very small nor very large).
Informally, this is a requirement of local typicalness;
e.g., $k=\lfloor N/2\rfloor$ is untypical for a large $N$
since it is described in such a simple way
(given $N$, which the definition assumes).
This characterization implies that the difference between $\DeX$ and $\DeR$
can be as large as $\frac12\lb N$ on data sequences of length $N$,
but not larger.

In the binary case
the difference between IID and exchangeability appears small,
of the order of magnitude $O(\log N)$,
which is much less than the attainable upper bound of $N+o(N)$
on $\DeX$ and $\DeR$.
In the algorithmic theory of randomness,
coincidence to within a logarithmic term is often considered
as being sufficiently close to disregard the difference.

These statements are also true about the definitions
$\DpX$ and $\DpR$ in terms of p-values,
as these inequalities show:
\begin{equation}\label{eq:lea}
  \begin{aligned}
    \DeX &\lea \DpX \lea \DeX + 2\lb\DeX,\\
    \DeR &\lea \DpR \lea \DeR + 2\lb\DeR
  \end{aligned}
\end{equation}
(they can be proved in the same way as Proposition~1
in \cite{Novikov:arXiv1608}).
We can see that $\DeX$ and $\DpX$, as well as $\DeR$ and $\DpR$,
coincide to within logarithmic terms.
Therefore, $\DpX$ and $\DpR$ also coincide to within a logarithmic term.
This may be the reason why Kolmogorov used $\DpX$
rather than $\DpR$ as formalization, in the binary case,
of ``a result of independent tests with a probability $p$ of getting a one
during each test'' \cite[Sect.~2]{Kolmogorov:1968}.

\begin{remark}
  In \cite[Sect.~2]{Kolmogorov:1968} Kolmogorov says
  (in translation) about his proposed definition,
  ``We view, approximately, in this manner `Bernoulli sequences'
  where separate signs are `independent'
  and appear with a certain probability $p$.''
  It is natural to assume, which I did, that the word ``approximately'' here
  means that he is ignoring the $O(\log N)$ difference
  between the two deficiencies (IID vs exchangeability).
  However, Kolmogorov told me that this was not what he meant
  (and he did not elaborate further).
\end{remark}

From the vantage point of conformal prediction, however,
the difference of $\frac12\lb N$ between $\DeX$ and $\DeR$ is not small at all.
Before discussing this, let us check that this difference
persists when we move to the p-versions, $\DpX$ and $\DpR$.
Indeed, let us consider a data sequence $\zeta\in\{0,1\}^N$
which is a typical element of the set of all binary sequences of length $N$
containing exactly $\lfloor N/2\rfloor$ 1s, for a large $N$.
Then $\DpX(\zeta)$ will be close to 0,
while, by the local limit theorem \cite[Sect.~1.6]{Shiryaev:2016},
$\DpR(\zeta)$ will be close to $\frac12\lb N$.
Therefore, the difference between $\DpX$ and $\DpR$
can also be as large as $\frac12\lb N$.

An ideal picture of conformal prediction will be introduced in the next section,
but what is important for us now is that the largest p-deficiency
at which an ideal conformal predictor can reject a false label for a test object is $\lb N$,
where $N$ is the length of the ``augmented training sequence''
(for details, see \eqref{eq:limitation} below).
Another manifestation of this phenomenon,
which we will call the ``fundamental limitation of conformal prediction'',
is the fact that the smallest possible conformal p-value is $1/N$.
Now $\frac12\lb N$ does not look small any more.
Even in the binary case, the difference between $\DpX$ and $\DpR$
can eat up half of the largest p-deficiency
achievable by an ideal conformal predictor.
In the non-binary case, the gulf between IID and exchangeability
becomes huge;
see inequality \eqref{eq:difference} below and its discussion.

The paper \cite{Vovk:1986-local} did not contain any proofs.
Full proofs were first published only in 2016,
but the main components appeared in \cite{Vovk:1997-local},
as indicated in \cite[Appendix~C]{Vovk:1986-local}.

\begin{remark}
  There are versions of de Finetti's theorem for finite sequences
  that assert near equivalence between a finite sequence being IID
  and being a prefix of
  a much longer finite sequence that is exchangeable
  \cite{Diaconis/Freedman:1980finite}.
  This idea of using exchangeable extensions makes it possible
  to adapt de Finetti's theorem to finite sequences,
  but here we will be only interested in basic exchangeability,
  with a fixed length of the data sequence.
\end{remark}

\section{Meeting Alex \iftoggle{TR}{Gammerman }{}and \iftoggle{TR}{Vladimir }{}Vapnik;
  emergence of conformal prediction}
\label{sec:Alex}

I first met Alex in Barcelona at EuroCOLT 1995,
the Second European Conference on Computational Learning Theory
(Barcelona, Spain, 13--15 March 1995).
Shortly before we met, Norman Gowar,
the Principal of Royal Holloway, University of London,
had suggested that Alex becomes the next Head of Department of Computer Science,
and Alex agreed, despite some misgivings.
Alex suggested creating a machine-learning group in the department,
and the Principal enthusiastically agreed.
One of Alex's goals in attending EuroCOLT 1995 was
to meet, as new Head of Department, active researchers in machine learning.

It is interesting that the First European Conference on Computational Learning Theory
was held at Royal Holloway, University of London, on 20--22 December 1993,
but neither Alex nor I attended it
(even though Alex had started teaching at Royal Holloway in September 1993).

Apart from discussing research at EuroCOLT 1995
(in particular, I learned about Alex's interest in Kolmogorov complexity),
I remember an enjoyable walk past the Columbus monument
at the bottom of La Rambla, the main pedestrian street in Barcelona.
My paper presented at the conference
(and published in the proceedings as \cite{Vovk:1995-local})
elaborated on the key element of the connection between IID and exchangeability
found in \cite{Vovk:1986-local}
(I talked about it at length in the previous section).
Later Paul Vit\'anyi invited me to submit an extended version of \cite{Vovk:1995-local}
to a Special Issue of the \emph{Journal of Computer and System Sciences}
devoted to EuroCOLT 1995;
the extended journal version appeared as \cite{Vovk:1997-local}
and later led to the publication of the proofs in \cite{Vovk:1986-local}.

In the summer of 1995 I moved to Stanford
to spend a year at the Centre for Advanced Studies in the Behavioral Sciences
(now part of Stanford University but then an independent institution).
It had been difficult to survive doing science in Russia
(I even attended a Business School in Moscow for a year,
with an internship in the USA in the summer of 1992),
and when Alex invited me to apply for a lectureship position at Royal Holloway
to join their machine-learning group
(later called CLRC, Computer Learning Research Centre),
I thought it was an exciting opportunity.
In December 1996 I had an interview there,
and at the same time my friend Philip Dawid arranged an interview at UCL
in case I was not successful at Royal Holloway.
But I was
(and have no idea of how successful I was at the UCL interview).
Even though I was hired as lecturer,
Alex told me that I would be promoted
to Professorship within three years (and I was).

Alex's first two hires were Vladimir Vapnik (part-time) and, shortly afterwards, I.
My family and I moved permanently to the UK in June 1996,
and that summer Alex, Vapnik, and I had very fruitful discussions
which later led, among other things, to the development of conformal prediction.
Vapnik was working (mainly or even exclusively)
on support vector machines and writing his 1998 book \cite{Vapnik:1998},
we discussed them repeatedly,
and I was eager to contribute.

Before meeting Vapnik, I had not taken the IID assumption seriously.
My philosophy was affected by my work
on what Shafer and I later called
``game-theoretic probability'' (\cite{Vovk:1993logic},
with later books \cite{Shafer/Vovk:2001,Shafer/Vovk:2019} joint with Shafer),
and as I mentioned earlier, this assumption appeared very narrow to me.
But Vapnik was taking it very seriously
and in many cases did not even mention explicitly that he was making it
(which at first even made it difficult for me to understand what he was saying).
It was a live demonstration of its importance,
and indeed I soon realized that it was the most fundamental assumption in machine learning.
And the problem of prediction under the IID assumption
looked much more down-to-earth and less philosophical
than providing frequentist foundations of probability
(my preferred approach to the foundations of probability
being based on Kolmogorov's Principle~B rather than Principle~A).

It was very natural to apply what I knew about typicalness deficiency
to Vapnik's IID picture,
as described briefly in \cite[Sect.~2.9.2]{Vovk/etal:2022-local}.
The ideal picture of prediction under IID or exchangeability
is straightforward (and described in \cite{Vovk/etal:1999}).
Let us suppose that each observation $z$ consists of two components,
an object $x$ and its label $y$,
and our task is to predict the label of a test object.
Suppose the observation space $\mathbf{Z}:=\mathbf{X}\times\mathbf{Y}$ is finite,
where $\mathbf{X}$ is the object space and $\mathbf{Y}$ is the label space,
both non-empty.
Let $\left|\mathbf{Y}\right|>1$.
The possibility of the decomposition $\mathbf{Z}:=\mathbf{X}\times\mathbf{Y}$
does not restrict generality
since we allow $\left|\mathbf{X}\right|=1$.
The upper or lower semicomputable functions producing IID and exchangeability deficiencies
are given both $\mathbf{X}$ and $\mathbf{Y}$ as inputs
(which are subsets of fixed aggregates of constructive objects).
Given a training sequence $z_1,\dots,z_n$ and a test object $x_{n+1}$,
our task is to predict the label $y_{n+1}$ of $x_{n+1}$.
We say that $y_{n+1}$ is the \emph{true label} of the test object $x_{n+1}$
while labels $y\ne y_{n+1}$ are \emph{false}.
The number $N:=n+1$ can be interpreted,
as is often done in conformal prediction,
as the length of the ``augmented training sequence''
(the training sequence extended by the test object $x_{n+1}$
with a possible label $y$).

In ``universal prediction'' we can use typicalness deficiency
for evaluating the plausibility of various potential labels
for the test object $x_{n+1}$.
For that we can use any of $\DpX$, $\DpR$, $\DeX$, or $\DeR$,
but for concreteness, let us concentrate on Kolmogorov's $\DpX$
(which is particularly close to conformal prediction).

\begin{remark}
  In the rest of this paper I will avoid using the expression ``universal prediction''
  because of another unfortunate terminological clash
  (in addition to that between statistical randomness and algorithmic randomness
  discussed in Sect.~\ref{sec:introduction}).
  On one hand, the adjective ``universal'' may mean
  ``related to universal partial computable functions'',
  as in ``universal p-test''.
  On the other hand, Nouretdinov et al.\ \cite{Nouretdinov/etal:2003ALT}
  applied it to conformal prediction meaning that, under IID,
  it does not lose much in efficiency
  as compared with any other prediction method that is valid in the same sense.
  The two senses are very different,
  so I will usually say ``ideal prediction'' rather than ``universal prediction''
  when talking about prediction in the ideal picture
  based on universal partial computable functions.
\end{remark}

Our prediction and how confident we can be in it can be figured out
by looking at the exchangeability deficiencies
\begin{equation}\label{eq:f}
  f(y)
  :=
  \DpX(z_1,\dots,z_n,x_{n+1},y)
\end{equation}
for various potential labels $y\in\mathbf{Y}$ for the test object.
(I am omitting parentheses in expressions
such as $\DpX(z_1,\dots,z_n,(x_{n+1},y))$
if this is unlikely to lead to a misunderstanding.)
For example, we can use
\[
  \hat y_{n+1}
  \in
  \arg\min_{y\in\mathbf{Y}}
  \DpX(z_1,\dots,z_n,x_{n+1},y)
\]
(let us assume, for simplicity, that the $\arg\min$ is attained at one point only)
as the \emph{point prediction} for the true test label $y_{n+1}$.
However, the full \emph{prediction function} $f$ defined by \eqref{eq:f}
contains a lot of other useful information.
For example, we can be confident that our point prediction is correct,
$\hat y_{n+1}=y_{n+1}$,
if the second smallest value $f(y)$ is large
(presumably the smallest value is $f(y_{n+1})$ under exchangeability).

The point prediction $\hat y_{n+1}$ complemented by the second smallest value of $f$
is a useful summary of the full prediction function $f$.
Another way to summarize the prediction function \eqref{eq:f}
is to fix a significance level $\epsilon\in\Q$
(such as $5\%$ or $1\%$)
and output a prediction set using $-\lb\epsilon$ as threshold,
\begin{equation}\label{eq:Gamma-log}
  \Gamma^{\epsilon}
  :=
  \{y\in\mathbf{Y}: \DpX(z_1,\dots,z_n,x_{n+1},y)<-\lb\epsilon\}
\end{equation}
(as in \cite[Sect.~2.2.4]{Vovk/etal:2022-local}
but with p-values measured on the logarithmic scale).
Notice that in this ideal picture the prediction sets are constructively closed
(i.e., their indicator functions are upper semicomputable),
which is natural:
when computing the ideal prediction sets
we keep making them narrower and narrower
(i.e., better and better)
as time passes.

The next question is how to make this ideal picture computable,
so that we could use, e.g., support vector machines
to find some practical approximations to the ideal prediction sets \eqref{eq:Gamma-log}.
This was a well-rehearsed step,
which I had done earlier in, e.g., \cite{Vovk:1993logic}
when developing game-theoretic probability.
The idea is to use the algorithmic theory of randomness
for getting a clear intuitive picture of some area of probability or statistics,
and then to strip the picture of its algorithmic content.
This makes results more precise and, in particular,
eliminates unspecified additive constants.
The process is described in detail in \cite[Sect.~6]{Vovk/Shafer:2003},
where Shafer and I present the algorithmic theory of randomness as a tool of discovery.

\begin{figure}[bt]
  \begin{center}
    \includegraphics[width=0.6\textwidth]{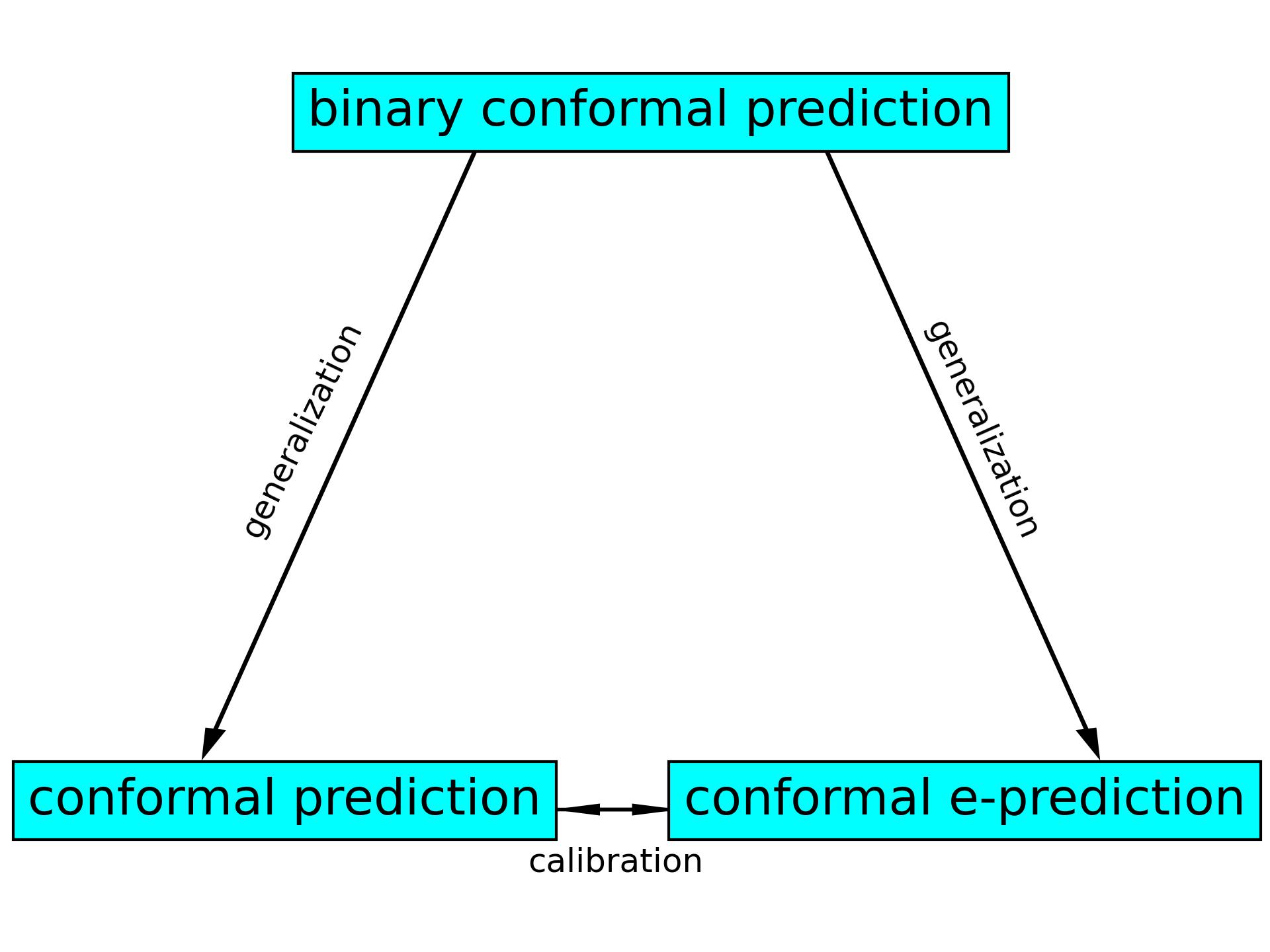}
  \end{center}
  \vspace{-0.5cm}
  \caption{Binary conformal prediction as special case
    of both conformal prediction and conformal e-prediction.
    Calibration will be discussed in Subsect.~\ref{subsec:other}.\label{fig:binary}}
\end{figure}

Conformal prediction in its primitive binary form,
which is a special case of both conformal prediction and conformal e-prediction,
was introduced in \cite{Gammerman/etal:1998}.
It was applicable only to binary classification,
since it was based on support vector machines.
The nonconformity measure used in that paper assigns
nonconformity scores of 1 to support vectors
and nonconformity scores of 0 to all other observations.
Let us call conformal prediction based on binary conformity measures
\emph{binary conformal prediction}.
See Fig.~\ref{fig:binary} for a pictorial representation,
and see \cite[Sect.~2]{Vovk/Wang:2023} for a more general comparison
of hypothesis testing based on p-values and e-values
with binary testing based on Cournot's principle.
The exposition in \cite{Gammerman/etal:1998}, however,
emphasized conformal e-prediction much more prominently
than conformal prediction.

\begin{remark}
  In \cite{Gammerman/etal:1998} we apply binary conformal prediction
  to a binary classification problem.
  This is a random coincidence, and binary conformal prediction
  is applicable to a wide range of prediction problems
  (see \cite[Sect.~3]{OCM44} for an example).
\end{remark}

Perhaps the first public announcement of conformal prediction
in a wide sense (namely, of binary conformal prediction)
happened in Alex's inaugural lecture in December 1997,
which was later published locally as \cite{Gammerman:1997-local}.
He and I considered it to be a public report about the work done at CLRC
and worked on it together;
my main contribution was to its section ``Transduction''
describing the binary conformal predictor based on support vector machines
and connecting it to Vapnik's idea of transductive inference.
Conformal prediction proper was introduced in \cite{Vovk/etal:1999}
(and soon afterwards in \cite{Saunders/etal:1999},
which mainly concentrated on support vector machines).

Untypically for literature on conformal prediction,
the paper \cite{Vovk/etal:1999} that introduced it paid some attention
to the ideal picture based on \eqref{eq:f}.
One remark that it makes \cite[Sect.~3, Remark~5]{Vovk/etal:1999} is that,
in the binary classification problem ($\mathbf{Y}=\{-1,1\}$),
if the true data sequence $z_1,\dots,z_N$,
where $N=n+1$ and $z_N=(x_N,y_N)$ is the true test observation,
is typical under IID,
then the maximal value of the prediction function \eqref{eq:f} will be $\lb N$.
This is again a manifestation of the fundamental limitation of conformal prediction
already mentioned in the previous section.

Two other results stated in \cite{Vovk/etal:1999}
extended relations between IID and exchangeability
discussed in the previous section
to the case of a general observation space $\mathbf{Z}$;
it turned out that Theorems~1 and~2 of \cite{Vovk:1986-local}
behave very differently when the assumption $\mathbf{Z}=\{0,1\}$ is dropped.
Theorem~\ref{thm:connection} carries over to the case of general $\mathbf{Z}$
without any problems.
On the other hand, a chasm between IID and exchangeability opens up
when $\mathbf{Z}$ is large
(or even infinite, which is allowed in \cite{Vovk/etal:1999});
namely,
\begin{equation}\label{eq:difference}
  \sup_{\zeta\in\mathbf{Z}^N}
  \DpR(\lbag\zeta\rbag)
  \gea
  \sup_{\zeta\in\mathbf{Z}^N}
  \DeR(\lbag\zeta\rbag)
  \gea
  N\lb\e-\frac12\lb N
\end{equation}%
\cite[Theorem~4]{Vovk/etal:1999},
where $\DpR(\lbag\cdot\rbag)$ is defined analogously to \eqref{eq:configuration}.
The difference between IID and exchangeability deficiencies now dwarfs
the best p-deficiency of $\lb N$ that can be used for prediction
(the fundamental limitation of conformal prediction;
cf.\ \eqref{eq:limitation} below).
Formally, we did not allow $\left|\mathbf{Z}\right|=\infty$
(just for simplicity of definitions),
but it is sufficient to assume that $\left|\mathbf{Z}\right|\ge N$.
There are no proofs in \cite{Vovk/etal:1999},
but the argument in the proof of \cite[Theorem~2, (16)]{Vovk:2020Yuri}
also proves~\eqref{eq:difference}.

In principle, the huge difference \eqref{eq:difference} per se
does not necessarily imply that IID and exchangeability are so very different:
\emph{a priori}, both can be huge, much greater than the difference.
However, we can complement \eqref{eq:difference} by
\begin{equation*}
  \sup_{\zeta\in\mathbf{Z}^N}
  \DpX(\lbag\zeta\rbag)
  \eqa
  \sup_{\zeta\in\mathbf{Z}^N}
  \DeX(\lbag\zeta\rbag)
  \eqa
  0,
\end{equation*}
where $\DpX(\lbag\cdots\rbag)$ and $\DeX(\lbag\cdots\rbag)$ are also defined
analogously to \eqref{eq:configuration}.
Therefore, exchangeability deficiency can be small
while IID deficiency is huge.
A specific example of a data sequence demonstrating this
is an algorithmically random permutation of $1,\dots,N$;
while it is perfectly exchangeable, it does not look IID at all:
given $N$, its IID p- and e-deficiency is
\[
  \lb\frac{N^N}{N!}
  \sim
  N\lb\e-\frac12\lb N.
\]

\section{Nouretdinov et al.'s discovery:
  universality of conformal prediction} % in the batch mode
\label{sec:NVG}

In the previous section we saw that the difference
between IID and exchangeability deficiencies can be huge.
% dwarfing differences relevant in conformal prediction.
Does it mean that, under the IID assumption,
we can achieve much more than what can be achieved by conformal prediction,
which only relies on exchangeability?
An important discovery by Nouretdinov, V'yugin,
and Alex \cite{Nouretdinov/etal:2003ALT}
was that conformal prediction is universal:
% (in the sense different from the use of ``universal''
% so far in this paper):
we do not lose much even under IID when using conformal prediction.
(I was among the authors of early versions of this paper,
but at some point Volodya V'yugin's exposition became too technical for me,
and I switched to other projects.
The final version of the paper is still very generous about my contribution.)

To discuss the universality of conformal prediction under IID,
it is useful to distinguish between two sides of our prediction problem.
For concreteness,
let us talk about IID p-deficiency $\DpR$.
\begin{itemize}
\item
  If the true data sequence $z_1,\dots,z_n,x_{n+1},y_{n+1}$ looks IID,
  i.e., \[\DpR(z_1,\dots,z_n,x_{n+1},y_{n+1})\] is small,
  we are in the situation of \emph{prediction proper};
  we can output $y_{n+1}$ as a confident prediction for the label of the test object $x_{n+1}$
  if \[\DpR(z_1,\dots,z_n,x_{n+1},y)\] is large for all false labels $y$.
\item
  If the true data sequence $z_1,\dots,z_n,x_{n+1},y_{n+1}$ does not look IID,
  i.e., $\DpR(z_1,\dots,z_n,x_{n+1},y_{n+1})$ is large,
  we are in the situation of \emph{anomaly detection};
  in this case all of $\DpR(z_1,\dots,z_n,x_{n+1},y)$, $y\in\mathbf{Y}$,
  can be expected to be large.
\end{itemize}
Nouretdinov et al.\ were interested in prediction proper,
which is the most natural setting of the prediction problem.
While the huge difference between IID and exchangeability
might well show in anomaly detection,
it does not have to show in prediction proper.
In this terminology,
the remark in \cite[Sect.~3, Remark~5]{Vovk/etal:1999}
mentioned in the previous section
says that, in the situation of prediction proper,
the maximal value of the prediction function is $\lb N$
(in the case of binary classification).
Now at least we have a rough coincidence of the upper bounds:
what can be achieved under IID (namely, $\lb N$)
can also be achieved already by conformal prediction
($\lb N$ is allowed by its fundamental limitation).
This coincidence hints at the universality of conformal prediction,
but Nouretdinov et al.\ paint a much fuller picture.

An e-test $E$ for exchangeability or IID
is said to be \emph{train-invariant} if,
for all $n$ and for all data sequences $(z_1,\dots,z_n,z_{n+1})\in\mathbf{Z}^{n+1}$,
\begin{equation*}
  E(z_1,\dots,z_n,z_{n+1})
  =
  E(z_{\sigma(1)},\dots,z_{\sigma(n)},z_{n+1})
\end{equation*}
for all permutations $\sigma$ of $\{1,\dots,n\}$.
In this definition, $z_1,\dots,z_n$ is interpreted as training sequence
and $z_{n+1}$ as test observation.
If such an $E$ is used as predictor,
we can now refer to $\lbag z_1,\dots,z_n\rbag$ as \emph{training bag},
or colloquially as \emph{training set},
which is a standard expression in machine learning;
$E$ does not depend on the ordering of the bag.
In the same way we define train-invariant p-tests for exchangeability.
(Nouretdinov et al.\ used the expression ``invariant''
for our ``train-invariant'',
but in this paper we will use ``invariant'' in a different,
much narrower, sense.)

The first result reported in \cite{Nouretdinov/etal:2003ALT},
their Proposition~1,
is Ilia Nouretdinov's observation that the class of conformal predictors
(understood to be functions producing conformal p-values
for all possible labels $y\in\mathbf{Y}$ for the test object)
essentially coincides with the class of train-invariant p-tests for exchangeability.
Namely, each function in the former class is dominated
(in the sense of being less than or equal to)
by some function in the latter class,
and vice versa.

It is also easy to check that the class of train-invariant e-tests for exchangeability
essentially coincides, in the same sense, with the class of conformal e-predictors
as defined in \cite{Vovk:2025PR}.
(The only difference is that ``dominates''
means ``is greater than or equal to'' in the case of e-tests.)

There exist a universal train-invariant p-test for exchangeability,
a universal train-invariant e-test for exchangeability,
a universal train-invariant p-test for IID,
and a universal train-invariant e-test for IID.
We fix them and denote their binary logarithms
(with the sign reversed in the case of the p-tests)
by $\DptX$, $\DetX$, $\DptR$, and $\DetR$, respectively.

\begin{remark}
  Nouretdinov et al.\ \cite[Sect.~4.2]{Nouretdinov/etal:2003ALT}
  used the expression ``i-test'' rather than ``e-test'',
  and I had used ``i-values'' earlier in \cite[Sect.~5]{Vovk:2001Denmark}.
  When working on \cite{Vovk/Wang:2021}, I misremembered ``i-'' as ``e-''.
  In hindsight, ``e-'' (standing for ``expectation'')
  appears to be a better counterpart of ``p-''
  (standing for ``probability'' in the context of p-values)
  than ``i-'' (standing for ``integral'').
\end{remark}

The following theorem is the main result of \cite{Nouretdinov/etal:2003ALT}
(Theorem~2, slightly simplified).
In it, $n$ ranges over $\N$,
$\mathbf{Z}=\mathbf{X}\times\mathbf{Y}$ as described earlier,
$(z_1,\dots,z_n)$ (training sequence) ranges over $\mathbf{Z}^n$,
$(x_{n+1},y_{n+1})$ (test observation) over $\mathbf{Z}$,
and $y$ (possible labels of $x_{n+1}$) over $\mathbf{Y}$.
\begin{theorem}\label{thm:overall}
  Letting $\zeta:=(z_1,\dots,z_n)$ stand for the training sequence,
  we have
  \begin{multline}\label{eq:overall-1}
    \DpR(\zeta,x_{n+1},y)
    -
    2\lb\DpR(\zeta,x_{n+1},y)
    -
    4\DpR(\zeta,x_{n+1},y_{n+1})
    -
    4\lb\left|\mathbf{Y}\right|\\
    \lea
    \DptX(\zeta,x_{n+1},y)
    \lea
    \DpR(\zeta,x_{n+1},y).
  \end{multline}
\end{theorem}

Theorem~\ref{thm:overall} is Nouretdinov et al.'s statement of universality
for conformal prediction in classification problems.
By classification I mean, informally,
prediction with a small number $\left|\mathbf{Y}\right|$ of classes.
In this case and in the situation of prediction proper
(i.e., $\DpR(\zeta,x_{n+1},y_{n+1})$ also being small),
\eqref{eq:overall-1} implies that
\[
  \DptX(\zeta,x_{n+1},y)
  \approx
  \DpR(\zeta,x_{n+1},y),
\]
i.e., ideal conformal prediction is almost as efficient as ideal prediction under IID.

Since $\DpR\approx\DeR$ and $\DptX\approx\DetX$
(with the approximate equalities holding to within logarithmic terms),
\eqref{eq:overall-1} also holds,
perhaps with the coefficients $2$ and $4$ replaced by larger ones,
for $\DeR$ and $\DetX$ in place of $\DpR$ and $\DptX$.
In other words, conformal e-prediction is also universal
in classification problems.

Theorem~\ref{thm:connection} connecting IID and exchangeability
was an important component of the proof of Theorem~\ref{thm:overall}
in \cite{Nouretdinov/etal:2003ALT}
(that component was stated there as Proposition~7).
The role of this connection will be clearly seen
in the functional version of Nouretdinov et al.'s result
stated in the following section.

Now we can discuss properly the fundamental limitation of conformal prediction
and its significance.
Because of the upper limit of $1/(n+1)$ on conformal p-values,
we have
\begin{equation}\label{eq:limitation}
  \DetX(\zeta,x_{n+1},y)
  \lea
  \DptX(\zeta,x_{n+1},y)
  \lea
  \log(n+1),
\end{equation}
where $\zeta:=(z_1,\dots,z_n)$.
In the situation of classification and prediction proper,
$\DpR(\zeta,x_{n+1},y_{n+1})\eqa0$,
\eqref{eq:overall-1} implies
\[
  \DpR(\zeta,x_{n+1},y)
  \lea
  \lb N + O(\lb\lb N).
\]
This continues to hold with $\DeR$ in place of $\DpR$.
Therefore, the fundamental limitation of conformal prediction
is also a limitation of prediction under IID
in general classification problems
(not necessarily binary classification).

\section{Perspective of the functional theory of randomness}
\label{sec:functional}

As already mentioned, most of the groundbreaking results
in \cite{Nouretdinov/etal:2003ALT} (all but Proposition~1)
involve unspecified constants.
The goal of this section is to explain
how the functional theory of randomness makes those results more practical:
instead of dealing with functions defined to within an additive constant,
now we are dealing with inclusions and other relations
between various function classes.
Very few proofs will be given,
and most of them can be found in % OCM 43
\cite{Vovk:arXiv2502} (which also covers the case of regression,
while this paper is constrained to classification,
similarly to \cite{Nouretdinov/etal:2003ALT}).
Otherwise, this section is more detailed and self-contained
than the previous ones.

In one respect,
the setting of this section is simpler than our setting so far;
since unspecified constants are gone,
the observation space $\mathbf{Z}$ and the length of the training sequence $n$
do not need to vary explicitly.
Our setting can also be made more general for free;
since the theory of algorithms is also gone,
now we just assume that the object space $\mathbf{X}$
is a non-empty measurable space.
However, we are still interested in the classification problem,
where the label space $\mathbf{Y}$ is finite with $\left|\mathbf{Y}\right|\ge2$
and equipped with the discrete $\sigma$-algebra.
The observation space $\mathbf{Z}=\mathbf{X}\times\mathbf{Y}$
is then also a measurable space.
In informal explanations, I will assume that $\mathbf{Y}$ is a small set,
such as in the case of binary classification $\left|\mathbf{Y}\right|=2$
(it might be a good idea for the reader to concentrate on this case, at least at first).
Now both $\mathbf{Z}$ and the length $n$ of the training sequence $z_1,\dots,z_n$
can be fixed throughout the section.
Given a new test object $x_{n+1}$,
our task is to predict $x_{n+1}$'s label $y_{n+1}$.
We will be interested in ``confidence predictors'',
i.e., algorithms for this prediction problem
producing valid measures of confidence,
such as p-values or e-values.
While the notion of confidence predictor is informal,
later I will give formal definitions of several classes of confidence predictors.

\subsection{Eight function classes}

To translate Nouretdinov et al.'s results into the functional theory of randomness,
it is useful to introduce eight function classes
representing eight kinds of confidence predictors based on three dichotomies:
\begin{itemize}
\item
  the assumption about the data-generating mechanism
  can be IID (R) or exchangeability (X);
\item
  with each potential label of a test object we can associate its p-value
  or e-value;
\item
  optionally, we can require the train-invariance (abbreviated to ``t'')
  of the confidence predictor.
\end{itemize}
The combination X/p/t corresponds to the conformal predictors,
while the combinations R/p and R/e correspond to the most general confidence predictors
under the IID assumption.
Following \cite{Nouretdinov/etal:2003ALT},
one of our goals will be to establish the closeness of the conformal predictors
(i.e., X/p/t predictors)
to the R/p predictors;
this goal is attractive since confidence predictors based on p-values
enjoy a more intuitive property of validity than those based on e-values.
Our argument will also establish the closeness of the conformal e-predictors
(i.e., X/p/t predictors)
to the R/e predictors.
Such closeness can be interpreted as the universality
of conformal prediction and conformal e-prediction.
We will also consider simplified versions of these goals.

In the rest of this section we will explore
\begin{itemize}
\item
  the difference between IID and exchangeability predictors
  in Subsect.~\ref{subsec:Kolmogorov}
  (and our results in this subsection
  will be summarized in Corollary~\ref{cor:Kolmogorov}
  and simplified in Corollary~\ref{cor:e-universality}),
\item
  the effect of imposing the requirement of train-invariance
  % w.r.\ to the permutations of the training sequence
  in Subsect.~\ref{subsec:train-invariance}
  (summarized in Theorem~\ref{thm:t}),
\item
  and the difference between confidence predictors based on p-values
  and those based on e-values
  in Subsect.~\ref{subsec:other}.
\end{itemize}
The overall picture will be summarized in Corollary~\ref{cor:overall}
and simplified in Corollary~\ref{cor:p-universality},
both in Subsect.~\ref{subsec:other}.

My informal explanations will sometimes be couched in the language
of the ``naive theory of randomness''
postulating the existence of the largest or smallest,
as appropriate, element in each function class;
this element will be called ``universal''.
Even though formally self-contradictory,
this postulate makes some intuitive sense
along the lines of the algorithmic theory of randomness.
(To make statements of the naive theory of randomness more palatable,
it may sometimes be helpful to qualify them using words such as ``almost'',
but not in this paper.)
In this way,
instead of using the algorithmic theory of randomness
for both formal analysis and intuitive considerations,
we may use the functional theory of randomness for the former
and the naive theory of randomness for the latter.

\begin{figure}[bt]
  \vspace{0.5cm}
  \begin{center}
    \unitlength 0.50mm
    \begin{picture}(80,80)(-10,-10)  % the size of the box and the coordinates of the bottom left corner
      \thinlines
      \put(0,0){\line(1,0){70}}  % the bottom line (thin)
      \put(70,0){\line(0,1){70}} % the right line (thin)
      \put(0,70){\line(1,0){70}} % the top line (thin)
      \put(0,0){\line(0,1){70}}  % the left line (thin)
      \put(15,15){\line(1,0){40}} % the bottom interior line (thin)
      \put(55,15){\line(0,1){40}} % the right interior line (thin and thick)
      \put(15,55){\line(1,0){40}} % the top interior line (thin and thick)
      \put(15,15){\line(0,1){40}} % the left interior line (thin)
      \put(0,0){\line(1,1){15}}     % the SW slanted line (thin)
      \put(70,0){\line(-1,1){15}}   % the SE slanted line (thin and thick)
      \put(70,70){\line(-1,-1){15}} % the NE slanted line (thin)
      \put(0,70){\line(1,-1){15}}   % the NW slanted line (thin and thick)
      \thicklines % now thick lines
      \put(55,15){\red{\line(0,1){40}}} % the right interior line (thin and thick)
      \put(15,55){\red{\line(1,0){40}}} % the top interior line (thin and thick)
      \put(70,0){\red{\line(-1,1){15}}} % the SE slanted line (thin and thick)
      \put(0,70){\red{\line(1,-1){15}}} % the NW slanted line (thin and thick)
      % and now the notation for the classes; first external:
      \put(-4,74){\makebox(0,0)[cc]{$\PR$}}  % top left corner (most important)
      \put(-5,-5){\makebox(0,0)[cc]{$\PtR$}} % bottom left corner
      \put(75,-5){\makebox(0,0)[cc]{$\PtX$}} % bottom right corner
      \put(75,75){\makebox(0,0)[cc]{$\PX$}}  % top right corner
      % and now internal:
      \put(16,54){\makebox(0,0)[tl]{$\ER$}}  % top left corner
      \put(16,16){\makebox(0,0)[bl]{$\EtR$}} % bottom left corner
      \put(54,16){\makebox(0,0)[br]{$\EtX$}} % bottom right corner
      \put(54,54){\makebox(0,0)[tr]{$\EX$}}  % top right corner
    \end{picture}
    \qquad\qquad\qquad
    \unitlength 0.50mm
    \begin{picture}(80,80)(-10,-10)  % the size of the box and the coordinates of the bottom left corner
      \thinlines
      \put(64,0){\vector(-1,0){58}}  % the bottom embedding
      \put(70,5){\vector(0,1){60}}  % the right embedding
      \put(64,70){\vector(-1,0){58}} % the top embedding
      \put(0,5){\vector(0,1){60}}   % the left embedding
      \put(49,15){\vector(-1,0){28}} % the bottom interior embedding
      \put(55,20){\vector(0,1){30}}  % the right interior embedding
      \put(49,55){\vector(-1,0){28}} % the top interior embedding
      \put(15,20){\vector(0,1){30}}  % the left interior embedding
      \put(5,5){\line(1,1){5}}     % the SW slanted line (calibration)
      \put(65,5){\line(-1,1){5}}   % the SE slanted line (calibration)
      \put(65,65){\line(-1,-1){5}} % the NE slanted line (calibration)
      \put(5,65){\line(1,-1){5}}   % the NW slanted line (calibration)
      % and now the notation for the classes; first external:
      \put(-6,65){\framebox(12,10)[cc]{$\PR$}}  % top left corner
      \put(-6,-5){\framebox(12,10)[cc]{$\PtR$}}  % bottom left corner
      \put(64,-5){\framebox(12,10)[cc]{$\PtX$}} % bottom right corner
      \put(64,65){\framebox(12,10)[cc]{$\PX$}} % top right corner
      % and now internal:
      \put(9,50){\framebox(12,10)[cc]{$\ER$}}  % top left corner
      \put(9,10){\framebox(12,10)[cc]{$\EtR$}} % bottom left corner
      \put(49,10){\framebox(12,10)[cc]{$\EtX$}} % bottom right corner
      \put(49,50){\framebox(12,10)[cc]{$\EX$}}  % top right corner
    \end{picture}
  \end{center}
  \vspace{-0.2cm}
  \caption{A cube representing eight function classes.
    The polygonal chain $\PR$--$\ER$--$\EX$--$\EtX$--$\PtX$
    is shown in red in the left panel.\label{fig:classes}}
\end{figure}

Figure~\ref{fig:classes} shows the eight function classes as a cube in each panel.
Let us concentrate on its left panel for now ignoring the right one.
We start from the function class in the top left corner (of the exterior square), $\PR$.
It consists of all \emph{IID p-variables} on $\mathbf{Z}^{n+1}$, i.e.,
functions $P:\mathbf{Z}^{n+1}\to[0,1]$ such that,
for all IID probability measures $R=Q^{n+1}$ on $\mathbf{Z}^{n+1}$,
we have \eqref{eq:P}.
(By default all functions referred to as ``variables'' are assumed to be measurable.)
The importance of the class $\PR$ stems from IID
being the standard assumption of machine learning.

The p-variable $P$ can be used as a ``confidence transducer'',
in the terminology of \cite[Sect.~2.7.1]{Vovk/etal:2022-local}.
Given a training sequence $z_1,\dots,z_n$ and a test object $x_{n+1}$,
we can compute the p-value $P(z_1,\dots,z_n,x_{n+1},y)$
for each possible label $y$ for $x_{n+1}$
(as before, p-values and e-values are just values
taken by p-variables and e-variables, respectively).
We can regard $P(z_1,\dots,z_n,x_{n+1},\cdot)$ to be a fuzzy set predictor for $y_{n+1}$.
To obtain a crisp set predictor, we can choose a \emph{significance level} $\epsilon\in(0,1)$
and define the prediction set
\begin{equation}\label{eq:Gamma}
  \Gamma^{\epsilon}(z_1,\dots,z_n,x_{n+1})
  :=
  \left\{
    y\in\mathbf{Y}:
    P(z_1,\dots,z_n,x_{n+1},y)>\epsilon
  \right\}
\end{equation}
by thresholding (cf.~\eqref{eq:Gamma-log}).
By the definition of p-variables,
the probability of error (meaning $y_{n+1}\notin\Gamma^{\epsilon}(z_1,\dots,z_n,x_{n+1})$)
for this crisp set predictor will not exceed~$\epsilon$.

In \cite[Sect.~2.1.6]{Vovk/etal:2022-local} confidence predictors were defined
as nested families $\Gamma^{\epsilon}$, $\epsilon\in(0,1)$,
and were called \emph{conservatively valid}
if $\Gamma^{\epsilon}$ makes an error with probability at most $\epsilon$.
This includes the families defined by \eqref{eq:Gamma} as a subclass,
and in general the inclusion is proper.
However, the difference is not essential,
as spelled out in Propositions~2.14 and~2.15 of \cite{Vovk/etal:2022-local}.
We will refer to the IID p-variables $P:\mathbf{Z}^{n+1}\to[0,1]$
as \emph{IID p-predictors}
(or, more fully, IID confidence p-predictors).

The top right corner (``Kolmogorov's corner'')
in Fig.~\ref{fig:classes}, $\PX$,
is the class that consists of all \emph{exchangeability p-variables} on $\mathbf{Z}^{n+1}$, i.e.,
functions $P:\mathbf{Z}^{n+1}\to[0,1]$ satisfying \eqref{eq:P}
for all $\epsilon\in(0,1)$ and all exchangeable probability measures $R$ on $\mathbf{Z}^{n+1}$.
Such p-variables serve as \emph{exchangeability p-predictors}.
Naively, a data sequence $\zeta\in\mathbf{Z}^{n+1}$ is exchangeable (resp.\ IID)
if $U(\zeta)$ is not small,
$U$ being the universal exchangeability (resp.\ IID) p-variable.

The bottom left corner of Fig.~\ref{fig:classes}, $\PtR$,
is the class of the train-invariant IID p-variables (elements of $\PR$),
and its bottom right corner $\PtX$
is the class of the train-invariant exchangeability p-variables.

The top left corner $\ER$ of the interior square in Fig.~\ref{fig:classes}
consists of all \emph{IID e-variables} on $\mathbf{Z}^{n+1}$, i.e.,
functions $E:\mathbf{Z}^{n+1}\to[0,\infty]$ such that,
for all IID probability measures $R$ on $\mathbf{Z}^{n+1}$,
\begin{equation}\label{eq:E}
  \int E \dd R
  \le
  1
\end{equation}
(which generalizes \eqref{eq:E-test}).
By Markov's inequality, $1/\ER\subseteq\PR$
(where $1/\ER$ consists of all $1/E$, $E\in\ER$).
We will also refer to e-variables $E\in\ER$ as \emph{IID e-predictors}.
For a given training sequence $z_1,\dots,z_n$ and test object $x_{n+1}$,
the e-value $E(z_1,\dots,z_n,x_{n+1},y)$ for each $y\in\mathbf{Y}$
tells us how unlikely $y$ is as label $y$ for $x_{n+1}$.
Therefore, $E(z_1,\dots,z_n,x_{n+1},\cdot)$ is again a soft set predictor for $y_{n+1}$.

The other function classes in Fig.~\ref{fig:classes} are defined in a similar way;
$\EX$ consists of all \emph{exchangeability e-variables} on $\mathbf{Z}^{n+1}$, i.e.,
functions $E:\mathbf{Z}^{n+1}\to[0,\infty]$ satisfying \eqref{eq:E} for all exchangeable $R$.
Finally, $\EtR$ and $\EtX$ consist of all train-invariant
functions in $\ER$ and $\EX$, respectively.
As before, these e-variables may be referred to as e-predictors,
depending on context.
The right panel of Fig.~\ref{fig:classes}
shows all inclusions between our eight classes,
with an arrow $A\to B$ from $A$ to $B$ meaning $A\subseteq B$.

It is interesting that all four confidence predictors
on the right of the cubes in Fig.~\ref{fig:classes}
have names (either existing or trivial modifications of existing)
containing the word ``conformal'':
\begin{itemize}
\item
  $\PX$ (the exchangeability p-predictors) are the weak conformal predictors
  \cite[Sect.~2.2.8 and Proposition~2.9]{Vovk/etal:2022-local};
\item
  $\EX$ (the exchangeability e-predictors) are the weak conformal e-predictors;
\item
  $\PtX$ (the train-invariant exchangeability p-predictors) are the conformal predictors
  % \cite[Proposition~1]{Nouretdinov/etal:2003ALT}
  \cite[Proposition~2.9]{Vovk/etal:2022-local};
\item
  $\EtX$ (the train-invariant exchangeability e-predictors) are the conformal e-predictors
  (see Subsect.~\ref{subsec:train-invariance} below).
\end{itemize}
As already mentioned,
the equivalence of $\PtX$ and conformal prediction was first established
in \cite[Proposition~1]{Nouretdinov/etal:2003ALT}.

Returning to the very informal language of the naive theory of randomness,
our discussion of calibration in Subsect.~\ref{subsec:other}
will show that IID data sequences $\zeta\in\mathbf{Z}^{n+1}$ can be defined as those
for which $U(\zeta)$ is not large,
$U$ being the universal (i.e., largest in this context) IID e-variable.
Similarly, exchangeable data sequences $\zeta$ can be defined as those
for which $U(\zeta)$ is not large,
$U$ being the universal (largest) exchangeability e-variable.

Following \cite{Nouretdinov/etal:2003ALT},
we will connect two opposite vertices of the cube
in the left panel of Fig.~\ref{fig:classes},
$\PR$ (IID p-prediction) and $\PtX$ (conformal prediction).
These vertices are important
since $\PR$ corresponds to most general confidence prediction
under the standard assumption of machine learning
and $\PtX$ is understood very well and has been widely implemented
(see, e.g., \cite{Bostrom:2024} and \cite{Cordier:2023}).

A convenient path connecting $\PR$ and $\PtX$ is shown as the bold red polygonal chain
in the left panel of Fig.~\ref{fig:classes}.
It will be used in stating the closeness of $\PR$ and $\PtX$ considered as predictors
(Corollary~\ref{cor:overall} below, analogous to Nouretdinov et al.'s main result).
Namely, we will establish the closeness for each step in the path separately:
\begin{itemize}
\item
  The step from $\PR$ to $\ER$ (from p-values to e-values for IID)
  is the calibration step, to be discussed in Subsect.~\ref{subsec:other}.
\item
  The step from $\ER$ to $\EX$ (from IID to exchangeability) is the key one;
  we will call it \emph{Kolmogorov's step}.
  It is the topic of Subsect.~\ref{subsec:Kolmogorov}.
\item
  The step from $\EX$ to $\EtX$ (adding train-invariance) is easier
  (if we do not worry about its optimality).
  We will call it the \emph{train-invariance step}.
  It is discussed in Subsect.~\ref{subsec:train-invariance}.
\item
  The step from $\EtX$ (conformal e-prediction) to $\PtX$ (conformal prediction)
  is the e-to-p calibration step,
  and it is also one of the topics of Subsect.~\ref{subsec:other}.
\end{itemize}

However, we will also be interested in other edges
of the cube in the left panel of Fig.~\ref{fig:classes},
first of all the edge connecting $\PtR$ and $\PtX$.
A useful connection between these two classes will be obtained
as a by-product (Corollary~\ref{cor:p-universality}).

\subsection{Kolmogorov's step}
\label{subsec:Kolmogorov}

In principle, besides the eight function classes shown in Fig.~\ref{fig:classes},
we are also interested in the following two:
\begin{itemize}
\item
  the class $\EiR$ consisting of \emph{invariant IID e-variables}:
  $E\in\EiR$ if $E\in\ER$
  and $E$ is invariant w.r.\ to all permutations of its arguments
\item
  the analogous class $\PiR$ consisting of \emph{invariant IID p-variables},
  which is the class of all $P\in\PR$
  that are invariant w.r.\ to all permutations of their arguments.
\end{itemize}
In this paper we will only use $\EiR$.
When $E\in\EiR$ is chosen in advance and $E(z_1,\dots,z_N)$
is large for the realized data sequence $z_1,\dots,z_N$,
we are entitled to reject the hypothesis
that its configuration $\lbag z_1,\dots,z_N\rbag$ was generated in the IID fashion.
Therefore, $\EiR$ is the analogue of $\DeR(\lbag\cdot\rbag)$
in the functional theory of randomness
(and $\PiR$ is the analogue of $\DpR(\lbag\cdot\rbag)$).

The following theorem is the functional version of the relation \eqref{eq:1986}
between IID and exchangeability.

\begin{theorem}\label{thm:decomposition}
  The class $\ER$ is the pointwise product of $\EX$ and $\EiR$:
  \begin{equation}\label{eq:decomposition}
    \ER
    =
    \EX\EiR.
  \end{equation}
\end{theorem}

\noindent
The pointwise product of function classes $\EEE_1$ and $\EEE_2$
is defined as the class of all products $E_1E_2$ for $E_1\in\EEE_1$ and $E_2\in\EEE_2$,
where $E_1E_2$ is the pointwise product of functions,
$(E_1E_2)(\zeta):=E_1(\zeta)E_2(\zeta)$.
For a proof of Theorem~\ref{thm:decomposition},
see \cite[Corollary~3]{Vovk:2020Yuri}.
However, since this result is derived in \cite{Vovk:2020Yuri}
as a corollary of a much more general statement
\cite[Theorem~1]{Vovk:2020Yuri}
and in order to make the exposition more self-contained,
let me give a simple independent derivation.

\begin{myproof}[of Theorem~\ref{thm:decomposition}]
  We consider the probability space $\mathbf{Z}^{n+1}$
  equipped with an IID probability measure $R$
  and consider $Z_i$, $i=1,\dots,n+1$, to be $z_i$ regarded as a random observation
  (formally, $Z_i$ is the random element on that probability space
  defined by $Z_i(z_1,\dots,z_{n+1}):=z_i$).

  To prove the inclusion ``$\subseteq$'' in \eqref{eq:decomposition},
  let $E\in\ER$.
  Set
  \begin{align}
    F(z_1,\dots,z_{n+1})
    &:=
    \frac{1}{(n+1)!}
    \sum_{\pi}
    E(z_{\pi(1)},\dots,z_{\pi(n+1)}),
    \notag\\
    E'(z_1,\dots,z_{n+1})
    &:=
    \frac{E(z_1,\dots,z_{n+1})}{F(z_1,\dots,z_{n+1})}
    \notag
  \end{align}
  (with $0/0:=1$),
  $\pi$ ranging over the permutations of $\{1,\dots,n+1\}$.
  It is obvious that $E'\in\EX$,
  and it is also easy to check that $F\in\EiR$:
  \begin{multline*}
    \E(F(Z_1,\dots,Z_{n+1}))
    =
    \frac{1}{(n+1)!}
    \sum_{\pi}
    \E(E(Z_{\pi(1)},\dots,Z_{\pi(n+1)}))\\
    \le
    \frac{1}{(n+1)!}
    \sum_{\pi} 1
    =
    1
  \end{multline*}
  (the inequality uses the fact that $Z_{\pi(1)},\dots,Z_{\pi(n+1)}$ are IID).

  To prove the inclusion ``$\supseteq$'' in \eqref{eq:decomposition},
  let $E\in\EX$ and $F\in\EiR$.
  Let us check that their product is in $\ER$:
  \begin{align*}
    \E&
    \left(
      E(Z_1,\dots,Z_{n+1})
      F(Z_1,\dots,Z_{n+1})
    \right)\\
    &=
    \E
    \left(
      \E
      \left(
        E(Z_1,\dots,Z_{n+1})
        F(Z_1,\dots,Z_{n+1})
        \mid
        \mathcal{G}
      \right)
    \right)\\
    &=
    \E
    \left(
      F(Z_1,\dots,Z_{n+1})
      \E
      \left(
        E(Z_1,\dots,Z_{n+1})
        \mid
        \mathcal{G}
      \right)
    \right)
    \iftoggle{TR}{\\&}{}
    \le
    \E
    \left(
      F(Z_1,\dots,Z_{n+1})
    \right)
    \le
    1,
  \end{align*}
  where $\mathcal{G}$ is the bag $\sigma$-algebra
  as defined in \cite[Sect.~A.5.2]{Vovk/etal:2022-local};
  the first inequality follows from \cite[Lemma~A.3]{Vovk/etal:2022-local}.
\end{myproof}

In terms of the naive theory of randomness,
\eqref{eq:decomposition} implies that
the universal IID e-variable is the product
of the universal exchangeability e-variable
and the universal invariant IID e-variable.
This shows that % according to Theorem~\ref{thm:decomposition},
the difference between being IID and exchangeability
lies in the configuration being IID.
Therefore, the following theorem establishes
a connection between IID e-predictors and exchangeability e-predictors.

\begin{theorem}\label{thm:Kolmogorov}
  For each invariant IID e-variable $F$
  there exists an IID e-variable $G$ such that,
  for all $z_1,\dots,z_n$, $z_{n+1}=(x_{n+1},y_{n+1})$, and $y\ne y_{n+1}$,
  \begin{equation}\label{eq:Kolmogorov}
    G(z_1,\dots,z_n,z_{n+1})
    \ge
    \frac{1}{\e(\left|\mathbf{Y}\right|-1)}
    F(z_1,\dots,z_n,x_{n+1},y).
  \end{equation}
\end{theorem}

We apply this theorem
(see Corollary~\ref{cor:Kolmogorov} below)
in the context where
$z_1,\dots,z_n,z_{n+1}$ is the true data sequence
with $z_{n+1}$ being the test observation
and $y$ is a false label;
ideally such $y$ should be excluded by our confidence predictor.
If $z_1,\dots,z_{n+1}$ is IID and $F$ is the universal invariant IID e-variable,
$F(z_1,\dots,x_{n+1},y)$ will be small,
and so there will be little difference between the degrees to which
a false label for the test object will be rejected by the universal e-predictors
under IID and exchangeability.

Let me give an informal argument
why $\lbag z_1,\dots,z_n,x_{n+1},y\rbag$ not being IID for $y\ne y_{n+1}$
implies the true data sequence
\begin{widish}
  (z_1,\dots,z_n,x_{n+1},y_{n+1})
\end{widish}
not being IID either.
Consider, for simplicity, the case of binary labels.
If after flipping the last label in the true data sequence
$(z_1,\dots,z_n,x_{n+1},y_{n+1})$
the bag of its elements becomes non-IID,
then either already the original bag $\lbag z_1,\dots,z_n,x_{n+1},y_{n+1}\rbag$ was non-IID
or the last element $(x_{n+1},y_{n+1})$ was special in the true data sequence,
and in any case already the original data sequence was non-IID.
A formal proof is given in \cite{Vovk:arXiv2502}.

The following asymptotic result says
that the $\left|\mathbf{Y}\right|$ in the denominator of \eqref{eq:Kolmogorov}
is in some sense optimal.

\begin{theorem}\label{thm:anti-Kolmogorov}
  For each constant $c>1$ the following statement holds true
  for a sufficiently large $\left|\mathbf{Y}\right|$ and a sufficiently large $n$.
  There exists an invariant IID e\-/variable $F$
  such that for each IID e-variable $G$
  there exist $z_1,\dots,z_n$, $z_{n+1}=(x_{n+1},y_{n+1})$, and $y\ne y_{n+1}$
  such that
  \begin{equation}
    \notag
    G(z_1,\dots,z_n,z_{n+1})
    <
    \frac{c}{\e\left|\mathbf{Y}\right|}
    F(z_1,\dots,z_n,x_{n+1},y).
  \end{equation}
\end{theorem}

Theorem~\ref{thm:anti-Kolmogorov} is proved in \cite{Vovk:arXiv2502}.
The idea of the proof can be explained informally
using the algorithmic theory of randomness
(or even more informally using the naive theory of randomness):
we can make the label $y$ in the bag $\lbag z_1,\dots,z_n,x_{n+1},y\rbag$
encode the bag $\lbag y_1,\dots,y_n\rbag$ of the other labels;
if we also make $y$ easily distinguishable from the other labels,
the value $F(z_1,\dots,z_n,x_{n+1},y)$
of the universal invariant IID e-variable
will be large.

Now we can state explicitly
a corollary of Theorems~\ref{thm:decomposition} and~\ref{thm:Kolmogorov}
that expresses the universality of weak conformal e-prediction.

\begin{corollary}\label{cor:Kolmogorov}
  For each IID e-predictor $E$
  there exist an exchangeability e\-/predictor $E'$ and an IID e-variable $G$ such that,
  for all $z_1,\dots,z_n$, $z_{n+1}=(x_{n+1},y_{n+1})$, and $y\ne y_{n+1}$,
  \begin{equation}\label{eq:cor-Kolmogorov}
    E'(z_1,\dots,z_n,x_{n+1},y)
    \ge
    \frac{1}{\e(\left|\mathbf{Y}\right|-1)}
    \frac{E(z_1,\dots,z_n,x_{n+1},y)}{G(z_1,\dots,z_n,z_{n+1})}.
  \end{equation}
\end{corollary}

The informal interpretation of \eqref{eq:cor-Kolmogorov}
is that, in classification,
every false label $y$ for the test object is excluded
by an exchangeability e-predictor
once it is excluded by an IID e-predictor,
unless the true data sequence $(z_1,\dots,z_{n+1})$ is not IID.
For this interpretation, there is no need
to take $E$, $E'$, and $G$ universal.

\medskip

\begin{myproof}[of Corollary~\ref{cor:Kolmogorov}]
  Let $E$ be an IID e-variable.
  By Theorem~\ref{thm:decomposition},
  there exist an exchangeability e-variable $E'$
  and an invariant IID e-variable $E''$
  such that
  \begin{equation}\label{eq:star-1}
    E(z_1,\dots,z_n,x_{n+1},y)
    =
    E'(z_1,\dots,z_n,x_{n+1},y)
    E''(z_1,\dots,z_n,x_{n+1},y)
  \end{equation}
  for all $z_1,\dots,z_n,x_{n+1},y$.
  By Theorem~\ref{thm:Kolmogorov} there exists an IID e-variable $G$ such that
  \begin{equation}\label{eq:star-2}
    G(z_1,\dots,z_n,z_{n+1})
    \ge
    \frac{1}{\e(\left|\mathbf{Y}\right|-1)}
    E''(z_1,\dots,z_n,x_{n+1},y)
  \end{equation}
  for all $z_1,\dots,z_n$, $z_{n+1}=(x_{n+1},y_{n+1})$, and $y\ne y_{n+1}$.
  It remains to combine \eqref{eq:star-1} and~\eqref{eq:star-2}.
\end{myproof}

\begin{remark}\rm
  In Corollary~\ref{cor:Kolmogorov} it is possible to have, in principle,
  0 in the denominator in \eqref{eq:cor-Kolmogorov}.
  Our interpretation of an inequality $A\ge c\frac{B}{C}$,
  where $A,c,B,C$ are all nonnegative,
  covering the possibility of $C=0$ is that it is equivalent, by definition,
  to $A C\ge c B$.
  Similar remarks can be made about other results,
  such as Theorem~\ref{thm:t} below.
\end{remark}

An important variation on Corollary~\ref{cor:Kolmogorov}
is where the original IID e-predictor $E$ is already train-invariant.
Under the IID assumption,
it seems useless to consider predictors that are not train-invariant,
and indeed the requirement of train-invariance follows
\cite[Sect.~2]{Vovk:arXiv2502}
from fundamental statistical principles,
such as the sufficiency principle and the invariance principle.
In this case the resulting predictor $E'$ will also be train-invariant
and, therefore, a conformal predictor.

\begin{corollary}\label{cor:e-universality}
  For each train-invariant IID e-predictor $E$
  there exist a conformal e\-/predictor $E'$ and an IID e-variable $G$ such that,
  for all $z_1,\dots,z_n$, $z_{n+1}=(x_{n+1},y_{n+1})$, and $y\ne y_{n+1}$,
  we have \eqref{eq:cor-Kolmogorov}.
\end{corollary}

\noindent
Corollary~\ref{cor:e-universality} is a statement of universality of conformal e-prediction
under train-invariance.

\subsection{Train-invariance step}
% From exchangeability to conformal e-prediction
\label{subsec:train-invariance}

Let us say that $G=G(z_1,\dots,z_n\mid z_{n+1})$
is a \emph{test-conditional exchangeability e-variable}
(given the test observation)
if
\[
  \forall(z_1,\dots,z_{n+1}):
  \frac{1}{n!}
  \sum_{\sigma}
  G
  \left(
    z_{\sigma(1)},\dots,z_{\sigma(n)} \mid z_{n+1}
  \right)
  \le
  1,
\]
$\sigma$ ranging over the permutations of $\{1,\dots,n\}$.
This property implies $G\in\EX$.
If $G=G(z_1,\dots,z_n\mid z_{n+1})$ is large
for the universal test-conditional exchangeability e-variable $G$,
the sequence $z_1,\dots,z_n$ is not exchangeable given $z_{n+1}$.
(And $G$ being large is a stronger property
than $z_1,\dots,z_{n+1}$ not being exchangeable.)

For any exchangeability e-predictor $E$,
define the corresponding train-invariant exchangeability e-predictor $\bar E$
by
\begin{equation}
  \notag
  \bar E(z_1,\dots,z_n,z_{n+1})
  :=
  \frac{1}{n!}
  \sum_{\sigma}
  E
  \left(
    z_{\sigma(1)},\dots,z_{\sigma(n)},z_{n+1}
  \right),
\end{equation}
$\sigma$ again ranging over the permutations of $\{1,\dots,n\}$.
This is the train-invariance step.
The following theorem says that $\bar E$ is almost as good as $E$
in our prediction problem unless $z_1,\dots,z_{n+1}$ is not exchangeable.

\begin{theorem}\label{thm:t}
  For each exchangeability e-predictor $E$
  there exists a test\-/conditional exchangeability e-variable $G$ such that,
  for all $z_1,\dots,z_n$, all $z_{n+1}=(x_{n+1},y_{n+1})$, and all $y\ne y_{n+1}$,
  \begin{equation}
    \notag
    \bar E(z_1,\dots,z_n,x_{n+1},y)
    \ge
    \frac{1}{\left|\mathbf{Y}\right|-1}
    \frac{E(z_1,\dots,z_n,x_{n+1},y)}{G(z_1,\dots,z_n\mid z_{n+1})}.
  \end{equation}
\end{theorem}

The intuition behind Theorem~\ref{thm:t}
is that each exchangeability e-predictor can be made train-invariant
without significant loss of efficiency in classification proper.
For a simple proof, see \cite{Vovk:arXiv2502}.

\subsection{Other steps}
\label{subsec:other}

In previous sections we discussed the two interior red steps
shown in the left panel of Fig.~\ref{fig:classes}.
Here we will discuss the two other red steps
and summarize the overall picture
obtaining a version of \cite[Theorem~2]{Nouretdinov/etal:2003ALT}
in the functional theory of randomness.
These two steps are analogues of the inequalities
\eqref{eq:lea} in the functional theory of randomness.

Conversion from p-values to e-values (\emph{calibration})
and vice versa (\emph{e-to-p calibration}) is understood very well:
see, e.g., \cite[Sect.~2]{Vovk/Wang:2021}.
E-to-p calibration is particularly simple:
there is one optimal e-to-p-calibrator,
$e\mapsto\min(1/e,1)$ \cite[Proposition~2.2]{Vovk/Wang:2021}.
As for calibration, a decreasing function $f:[0,1]\to[0,\infty]$ is a \emph{calibrator}
(transforms p-values into e-values)
if and only if $\int_0^1 f \le 1$ \cite[Proposition~2.1]{Vovk/Wang:2021}.
We will use the calibrator
\begin{equation}\label{eq:calibrator}
  f(p)
  :=
  \delta
  p^{\delta-1}
\end{equation}
for a fixed value $\delta\in(0,1)$.
If $\delta$ is small, $f(p)$ will be close to $1/p$
if we ignore the multiplicative constant
(as customary in the algorithmic theory of randomness).
Other popular calibrators are
\begin{equation*}
  f(p)
  :=
  \begin{cases}
    \infty & \text{if $p=0$}\\
    \kappa (1+\kappa)^\kappa p^{-1} (-\ln p)^{-1-\kappa} & \text{if $p\in(0,\exp(-1-\kappa)]$}\\
    0 & \text{if $p\in(\exp(-1-\kappa),1]$}
  \end{cases}
\end{equation*}
for a constant $\kappa>0$
(see \cite[Appendix~B]{Vovk/Wang:2021};
this calibrator is even closer to $1/p$ than \eqref{eq:calibrator} with a small $\delta$)
and Shafer's \cite[Sect.~3, (6)]{Shafer:2021} calibrator
\[
  f(p) := p^{-1/2}-1.
\]
The lines between the corresponding $\mathcal{P}$ and $\mathcal{E}$ vertices
in the right panel of Fig.~\ref{fig:classes}
stand for the possibility of calibration or e-to-p calibration
(similarly to the double-headed arrow in Fig.~\ref{fig:binary}).

The following result combines all the previous statements in this section.

\begin{corollary}\label{cor:overall}
  Let $\delta\in(0,1)$.
  For all $P\in\PR$ there exist $P'\in\PtX$ and $G\in\ER$ such that,
  for all observations $z_1,\dots,z_n$, $z_{n+1}=(x_{n+1},y_{n+1})$,
  and labels $y\ne y_{n+1}$,
  \begin{multline}\label{eq:overall-2}
    P'(z_1,\dots,z_n,x_{n+1},y)\\
    \le
    \frac{\e(\left|\mathbf{Y}\right|-1)^2}{\delta}
    G(z_1,\dots,z_{n+1})^2
    P(z_1,\dots,z_n,x_{n+1},y)
    ^{1-\delta}.
  \end{multline}
\end{corollary}

Corollary~\ref{cor:overall} reduces (as usual, imperfectly)
IID p-predictors to conformal predictors.
It says that in classification problems
every false label excluded by an IID p-predictor
is excluded by a conformal predictor (perhaps less strongly)
unless the true data sequence is non-IID.
It is an analogue of \cite[Theorem~2]{Nouretdinov/etal:2003ALT}.

\smallskip

\begin{myproof}[of Corollary~\ref{cor:overall}]
  Let $P\in\PR$ and $\delta\in(0,1)$.
  We will construct $P'\in\PtX$ and $G\in\ER$
  satisfying \eqref{eq:overall-2} in several steps.
  Since \eqref{eq:calibrator} is a calibrator,
  there is $E\in\ER$ satisfying
  \begin{equation}\label{eq:step-1}
    E(z_1,\dots,z_n,x_{n+1},y)
    \ge
    \delta
    P(z_1,\dots,z_n,x_{n+1},y)^{\delta-1}
  \end{equation}
  (in fact, with ``$=$'' in place of ``$\ge$'');
  here and in the rest of the proof we will leave
  ``for all $z_1,\dots,z_n$, $z_{n+1}=(x_{n+1},y_{n+1})$, and $y\ne y_{n+1}$''
  implicit.
  By Corollary~\ref{cor:Kolmogorov},
  there exist $E'\in\EX$ and $G_1\in\ER$ such that
  \begin{equation}\label{eq:step-2}
    E'(z_1,\dots,z_n,x_{n+1},y)
    \ge
    \frac{1}{\e(\left|\mathbf{Y}\right|-1)}
    \frac{E(z_1,\dots,z_n,x_{n+1},y)}{G_1(z_1,\dots,z_n,z_{n+1})}.
  \end{equation}
  By Theorem~\ref{thm:t},
  there exist $E''\in\EtX$ and $G_2\in\ER$ such that
  \begin{equation}\label{eq:step-3}
    E''(z_1,\dots,z_n,x_{n+1},y)
    \ge
    \frac{1}{\left|\mathbf{Y}\right|-1}
    \frac{E'(z_1,\dots,z_n,x_{n+1},y)}{G_2(z_1,\dots,z_n,z_{n+1})}.
  \end{equation}
  Finally, since $e\mapsto1/e$ is an e-to-p calibrator,
  there is $P'\in\PtX$ satisfying
  \begin{equation}\label{eq:step-4}
    P'(z_1,\dots,z_n,x_{n+1},y)
    \le
    1/E''(z_1,\dots,z_n,x_{n+1},y).
  \end{equation}
  It remains to combine \eqref{eq:step-1}--\eqref{eq:step-4}
  and set $G:=\sqrt{G_1G_2}$.
  (By the inequality between the geometric and arithmetic means,
  $G\in\ER$.)
\end{myproof}

Of course, Corollary~\ref{cor:overall} continues to hold
if the condition $P\in\PR$ is replaced by $P\in\PX$.
In this case, however, we can drop step~\eqref{eq:step-2}
and replace~\eqref{eq:overall-2} by
\begin{equation*}
% \begin{wideformula*}
  P'(z_1,\dots,z_n,x_{n+1},y)
  \le
  \frac{\left|\mathbf{Y}\right|-1}{\delta} % \NarrowLinebreak
  G(z_1,\dots,z_{n+1})
  % \left(
    P(z_1,\dots,z_n,x_{n+1},y)
  % \right)
  ^{1-\delta}.
\end{equation*}
% \end{wideformula*}
The most important case, however, is where $P\in\PtR$.
Now we can drop step~\eqref{eq:step-3},
as spelled out in the following corollary.

\begin{corollary}\label{cor:p-universality}
  Let $\delta\in(0,1)$.
  For each $P\in\PtR$ there exist $P'\in\PtX$ and $G\in\ER$ such that,
  for all $z_1,\dots,z_n$, $z_{n+1}=(x_{n+1},y_{n+1})$, and $y\ne y_{n+1}$,
  \begin{equation*}
    P'(z_1,\dots,z_n,x_{n+1},y)
    \le
    \frac{\e(\left|\mathbf{Y}\right|-1)}{\delta}
    G(z_1,\dots,z_{n+1})
    % \left(
      P(z_1,\dots,z_n,x_{n+1},y)
    % \right)
    ^{1-\delta}.
  \end{equation*}
\end{corollary}

Corollary~\ref{cor:p-universality} reduces train-invariant IID p-prediction
to conformal prediction without significant loss in efficiency.
Since the condition of train-invariance is so natural under the IID assumption,
Corollaries~\ref{cor:e-universality} and~\ref{cor:p-universality}
may be the most useful statements in this section.

\section{Conclusion}

This paper further develops the functional theory of randomness
in the direction of Nouretdinov, V'yugin, and Gammerman's work
on universality of conformal prediction.
While the statements of the functional theory of randomness
are somewhat less intuitive than those of the algorithmic theory of randomness,
they are more precise (do not involve unspecified constants)
and simpler in an important respect:
e.g., the analogue of \eqref{eq:decomposition} in the algorithmic theory of randomness,
\eqref{eq:1986},
involves the condition ``$\mid\DeR(\lbag\zeta\rbag)$'',
which disappears in \eqref{eq:decomposition}.
In the naive theory of randomness we could write, instead,
\[
  \DeR(\zeta)
  =
  \DeX(\zeta)
  +
  \DeR(\lbag\zeta\rbag).
\]
For the reader familiar with the algorithmic theory of complexity,
the condition ``$\mid\DeR(\lbag\zeta\rbag)$'' is analogous to the second entry of ``$K(x)$''
in the Levin--Chaitin formula
\[
  K(x,y)
  \eqa
  K(x) + K(y\mid x,K(x))
\]
\cite[Theorem~67]{Shen/etal:2017book}.
This and similar conditions can be removed in the functional theory of randomness
and functional theory of complexity
(the latter is introduced in \cite[Appendix]{Vovk:2020Yuri}).

In Sect.~\ref{sec:RE} I emphasized the narrowness of both assumptions,
IID and exchangeability, whereas this paper concentrates on what can be done under IID.
There have been many developments in conformal prediction beyond exchangeability,
such as those in \cite{Tibshirani/etal:2019,Gibbs/Candes:2021,Barber/etal:2023,Angelopoulos/etal:2023};
see also the review \cite[Chap.~7]{Angelopoulos/etal:arXiv2411}.
For all of them, exchangeability serves as a starting point.

\subsection*{Acknowledgments}

Many thanks to Ruodu Wang for his comments.
As always, the Stack Exchange \TeX--\LaTeX\ community have been ready to help.

\end{document}